\documentclass[10pt,journal,compsoc]{IEEEtran}
\usepackage{grffile}
\usepackage{amsmath,amssymb} 
\usepackage{algorithm}
\usepackage{algorithmic}
\usepackage[margin=0pt,font=small,labelfont=small,justification=justified,labelsep=period]{caption}
\usepackage{bm}
\usepackage{makecell}
\usepackage{subfigure}
\usepackage{multicol}  
\usepackage{multirow}
\usepackage{booktabs}
\usepackage{array}
\usepackage{mathtools}
\usepackage{bbm}
\usepackage{color}
\newcommand{\etal}{\textit{et al}. }
\newcommand{\ie}{\textit{i}.\textit{e}., }
\newcommand{\eg}{\textit{e}.\textit{g}., }
\newcommand{\norm}[1]{\left\lVert#1\right\rVert}
\newcommand{\dkl}[2]{D_{KL}(#1\Vert#2)}
\usepackage{diagbox}
\usepackage[pagebackref=true,breaklinks=true,letterpaper=true,colorlinks,citecolor=blue,linkcolor=blue,bookmarks=false]{hyperref}

\ifCLASSOPTIONcompsoc
\usepackage[nocompress]{cite}
\else
\usepackage{cite}
\fi  

\begin{document}
\title{Shaping Deep Feature Space towards\\Gaussian Mixture for Visual Classification}
\author{Weitao~Wan,~
	Jiansheng~Chen,~
	Cheng Yu,~
	Tong~Wu,~
	Yuanyi~Zhong,~
	and~Ming-Hsuan~Yang
	\IEEEcompsocitemizethanks{\IEEEcompsocthanksitem Weitao Wan, Jiansheng Chen, Cheng Yu and Tong Wu are with the Department of Electrical Engineering, Tsinghua University, Beijing, 100084. Email: \{wwt16$\vert$yuc18$\vert$wutong16\}@mails.tsinghua.edu.cn, jschenthu@mail.tsinghua.edu.cn
		\IEEEcompsocthanksitem Yuanyi Zhong is with the Department of Computer Science, University of at Urbana-Champaign, IL, 61820. His work was done when he was with Tsinghua University. Email: yuanyiz2@illinois.edu
		\IEEEcompsocthanksitem Ming-Hsuan Yang is with the Department of Computer Science and Engineering, University of California, Merced, CA, 95340. Email:mhyang@ucmerced.edu}
}


\markboth{IEEE Transactions on Pattern Analysis and Machine Intelligence}{}

\IEEEtitleabstractindextext{%
\begin{abstract}
The softmax cross-entropy loss function has been widely used to train deep models for various tasks.
In this work, we propose a Gaussian mixture (GM) loss function for deep neural networks for visual classification. 
Unlike the softmax cross-entropy loss, our method explicitly shapes the deep feature space towards a Gaussian Mixture distribution.
With a classification margin and a likelihood regularization, the GM loss facilitates both high classification performance and accurate modeling of the feature distribution.
The GM loss can be readily used to distinguish abnormal inputs, such as the adversarial examples, based on the discrepancy between feature distributions of the inputs and the training set.
		%
Furthermore, theoretical analysis shows that a symmetric feature space can be achieved by using the GM loss, which enables the models to perform robustly against adversarial attacks.
The proposed model can be implemented easily and efficiently without using extra trainable parameters.
Extensive evaluations demonstrate that the proposed method performs favorably not only on image classification but also on robust detection of adversarial examples generated by strong attacks 
under different threat models.
		%
		%
\end{abstract}
	

}

\maketitle

\IEEEdisplaynontitleabstractindextext

%
\IEEEpeerreviewmaketitle

\IEEEraisesectionheading{\section{Introduction}\label{sec:introduction}}
\begin{figure*}[t]
	\footnotesize
	\centering
	\renewcommand{\tabcolsep}{2pt} 
	\renewcommand{\arraystretch}{1} 
	\begin{center}
		\begin{tabular}{ccc}
\includegraphics[width=0.3\linewidth]{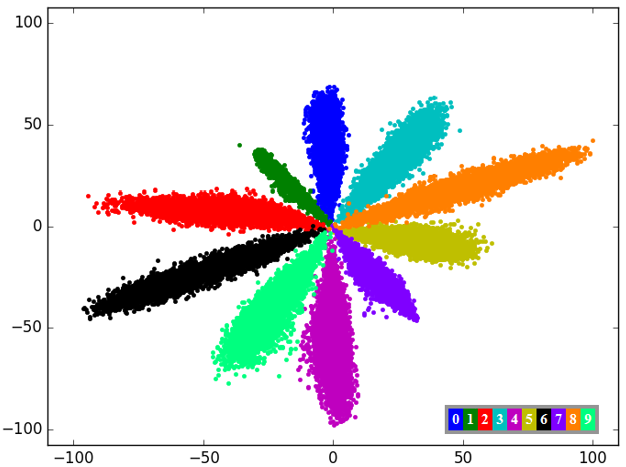}&		
\includegraphics[width=0.3\linewidth]{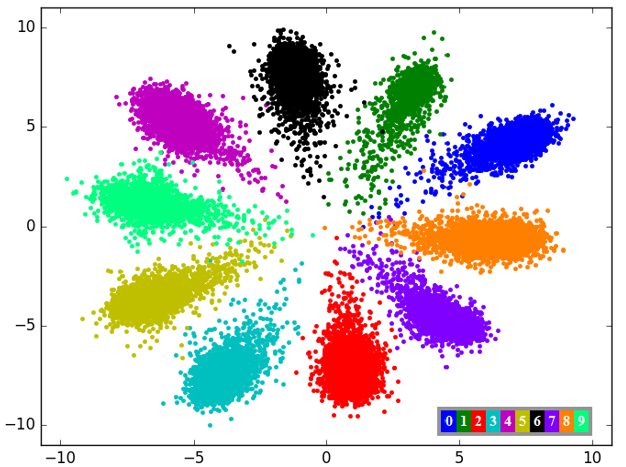}&
\includegraphics[width=0.3\linewidth]{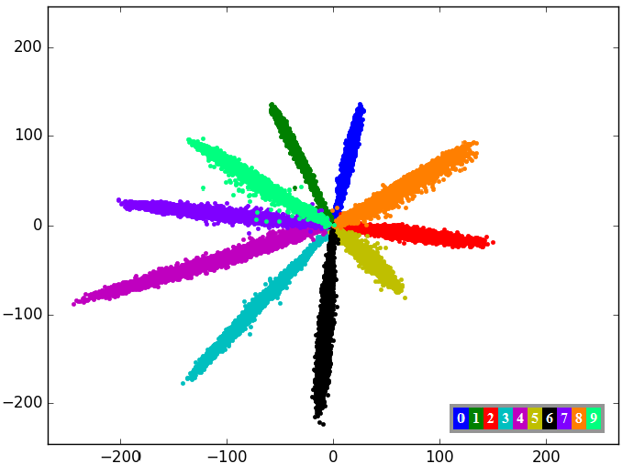}
 \\
(a) Softmax Loss &
(b) Center Loss~\cite{DBLP:conf/eccv/WenZL016}&
(c) L-Softmax Loss~\cite{DBLP:conf/icml/LiuWYY16}
\\
\hspace{0.3mm}
\includegraphics[width=0.3\linewidth]{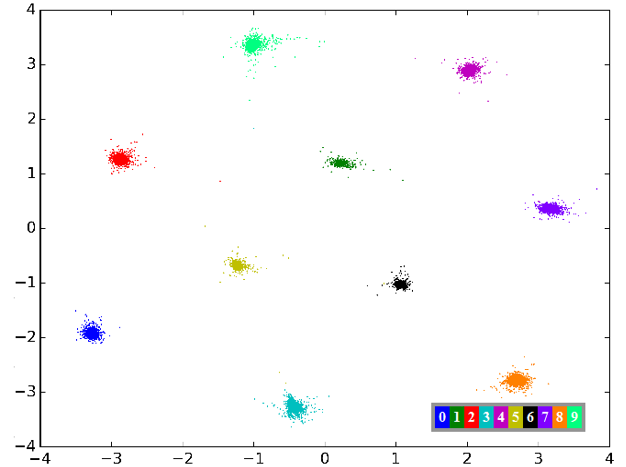}&
\hspace{0.01mm}
\includegraphics[width=0.3\linewidth]{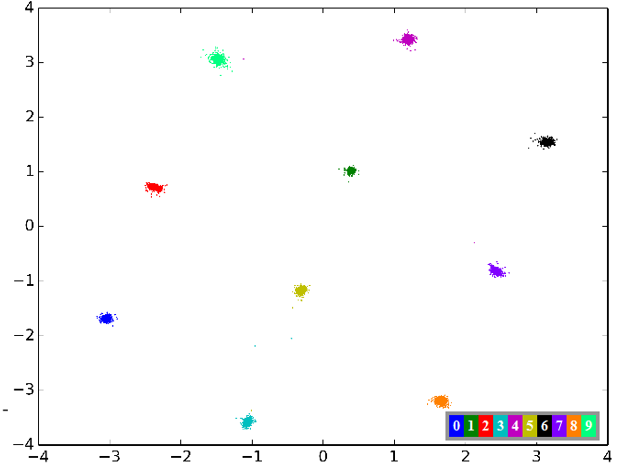}&
\hspace{1.49mm}
\includegraphics[width=0.3\linewidth]{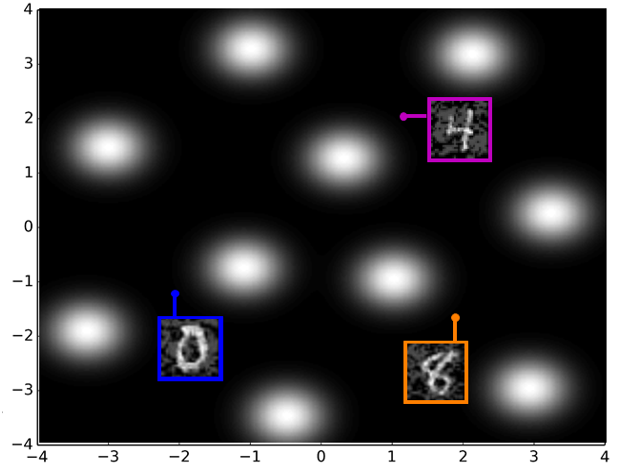}
\\

(d) GM loss w/o margin&
(e) GM loss w/ margin&
(f) Kernel Density Heatmap
\\	
		\end{tabular}
	\end{center}
	\caption{
		\textbf{Two-dimensional feature embeddings on the MNIST training set}. The Heatmap in (f) shows the kernel density values of features, in which brighter pixels indicate higher values. The adversarial examples generated by the Fast Gradient Sign Method \cite{goodfellow2014explaining} have extremely low kernel density values in the learned GM distribution and thus can be easily detected by our method.
	}
		\vspace*{-4mm}
\label{fig01}
\end{figure*}

\IEEEPARstart{D}{eep} neural networks have been applied to numerous tasks 
including object recognition \cite{DBLP:conf/nips/KrizhevskySH12, DBLP:conf/icml/IoffeS15, DBLP:conf/cvpr/HeZRS16}, face recognition \cite{DBLP:conf/cvpr/SchroffKP15, DBLP:conf/nips/SunCWT14} 
and speech recognition \cite{Dahl2011Context,Hinton2012Deep}, to name a few. 
For these tasks, the softmax cross-entropy loss (also known as the \emph{softmax loss}), has been widely adopted as the classification loss function for various deep neural networks \cite{DBLP:journals/corr/SzegedyVISW15, DBLP:conf/cvpr/HeZRS16, zagoruyko2016wide, larsson2016fractalnet, huang2017densely}.
For visual classification, the affinity score of an input sample to each class is first computed by a linear transformation on the extracted deep features. 
The posterior probability is then modeled as the normalized affinity scores using the softmax function.
Finally, the cross-entropy between the posterior probability and class label is used as the loss function.
The softmax loss has its probabilistic interpretation in that, for a large class of distributions, the posterior distribution complies with the softmax transformation of linear functions of the feature vectors \cite{christopher2006pattern}.
It can also be derived from a binary Markov Random Field or a Boltzmann Machine model \cite{deng2014large}.
However, the relationship between the affinity score and feature distribution is not entirely clear. 
In addition, for an extracted feature vector, its likelihood to the training feature distribution is not well formulated.

Numerous loss functions have been proposed to train deep models more effectively than those based on the vanilla softmax loss. 
The Euclidean distances between each pair \cite{DBLP:conf/nips/SunCWT14} or among each triplet \cite{DBLP:conf/cvpr/SchroffKP15} of extracted features are added as an additional loss to the softmax loss.
Alternatively,  the Euclidean distance between each feature vector and its class centroid is used \cite{DBLP:conf/eccv/WenZL016}. 
%
Within the formulation of the softmax loss, however, the similarity metric based on cosine distance is more effective since the affinity score is computed by inner product, 
indicating that using additional Euclidean distance based losses may not be an ideal option.
As such, an angular distance based margin is introduced in \cite{DBLP:conf/icml/LiuWYY16} to force extra intra-class compactness and inter-class separability, leading to better generalization of the trained models.
It has been shown that a classification margin is naturally embedded in the formulation of the softmax loss~\cite{kobayashiLarge}.
To leverage the naturally embedded margin, an extra regularization term has to be introduced to maximize the margin.
Nevertheless, the softmax loss is still indispensable and widely used in the training processes of these approaches.
Thus, the probabilistic model of the training feature space is still not well exploited.

Exploiting the distribution of deep features is of great interest for robust detection of adversarial examples in addition to classification accuracy.
For example, kernel density estimation methods have been used to discriminate between normal samples and adversarial examples~\cite{feinman2017detecting, pang2018rce}.
These approaches are more efficient and reliable than those based on distributions in the input space because the deep features contain richer semantic information.
However, existing schemes are developed based on pre-trained deep models for feature extraction and other modules to estimate distributions. 
%
%
Thus, it is of prime importance to develop an end-to-end method that explicitly shapes the deep feature distribution to facilitate both high classification performance and robustness against adversarial examples.
%

In this work, we propose a Gaussian Mixture loss (GM loss)  
to model deep features. 
As this work focuses on visual classification, the number of Gaussian components can be set to the number of object categories. 
As such, the posterior probability can be explicitly computed using the Bayes rule.
The classification loss is computed as the cross-entropy between the posterior probability and the corresponding class label.
However, adopting this classification loss alone for training is not sufficient to shape deep features towards a Gaussian mixture distribution since suboptimal solutions may be obtained in which the posterior probability for the ground-truth class is large but the features are far away from all the Gaussian centers.
In other words, optimizing the classification loss alone can facilitate the classification task but the feature distribution may significantly deviate from the expected Gaussian mixture.
To address this issue, we further introduce a novel regularization term to maximize the likelihood of deep features under the Gaussian mixture distribution, thereby enforcing the features to follow the assumed distribution.
As such, the proposed GM loss is a combination of the classification loss and the likelihood regularization.
By utilizing the GM loss for training through typical stochastic gradient descent, the deep models are capable of extracting deep features that are discriminative for classification while following the assumed distribution, which makes the GM loss intrinsically different from the softmax loss.
Furthermore, we present theoretical analysis to reveal that such a feature distribution is advantageous for the robustness against adversarial attacks.

The proposed regularization term essentially differs from the one in our preliminary results~\cite{gm}. 
In this work, we maximize the likelihood of deep features under the Gaussian mixture distribution instead of one Gaussian component of the ground-truth class~\cite{gm}.
However, directly adopting the likelihood function as the regularization term leads to a trivial solution, in which all the feature vectors either collapse to the origin or scale up such that the $L_2$ norm is as large as possible, depending on the loss weight for the regularization term.
To address this issue, we maximize the lower bound of the likelihood function which leads to a simpler formula and better equilibrium between the classification loss and the regularization term.
This regularization facilitates learning effective representation in several aspects. 
First, unlike the original form~\cite{gm}, it helps weight regularization by constraining the overall length of Gaussian means and feature vectors, achieving better generalization ability.
Second, we show that it leads to a symmetric optimal solution where the distances between each pair of Gaussian means are equal. 
Similar approaches have been shown to be effective for robustness against adversarial attacks
\cite{pang2018rce,pang2018max}.
Using the proposed loss to train deep models, the probability distribution of the features can be explicitly formulated for a well-trained model.
Fig.~\ref{fig01} shows that the learned feature distribution using the proposed GM loss is significantly different from that learned using the softmax loss or variants.

The formulation of the GM loss also facilitates incorporating a classification margin during training.
As a result, it is not necessary to use an additional complicated distance function as in the large-margin softmax loss \cite{DBLP:conf/icml/LiuWYY16} or add an extra loss term to leverage the margin as in \cite{kobayashiLarge}.
The proposed classification margin is designed to adapt to the current distance between features and the Gaussian means of ground-truth classes.
It is more effective than a constant margin adopted in the triplet loss~\cite{schroff2015facenet} as the proposed margin imposes a larger penalty on the samples which the model has not learned well yet.
%
%
%
In addition, it enjoys several merits than the softmax loss and variants. 
First, the GM loss helps to achieve better classification performance by incorporating a classification margin in an efficient manner.
The improvement can be obtained for various CNN architectures on different datasets including large-scale ones such as ImageNet~\cite{imagenet}.
Second, by shaping the deep features towards a Gaussian mixture distribution, the GM loss leads to the feature space in which features of abnormal inputs have low likelihood in the learned feature distribution.
Thus, it is effective to discriminate adversarial inputs from clean ones based on the discrepancy of feature distributions.
With some approximations, it can be theoretically proven that the GM loss guarantees a symmetric feature space, which has been shown in prior work~\cite{pang2018rce, pang2018max} to be effective for defending against adversarial attacks.

The main contributions of this work are:
\begin{itemize}
	\item We propose  the Gaussian Mixture loss to train Convolutional Neural Networks for visual classification.
	In contrast to the commonly used softmax loss or variants, the GM loss can explicitly model the feature distribution.
	This formulation facilitates incorporating a classification margin in an effective manner.
	\item By introducing a likelihood regularization, the deep features are encouraged to follow a Gaussian Mixture distribution. 
	We show that by incorporating the regularization, the GM loss theoretically leads to a symmetric feature distribution in which the distances between any two Gaussian means are equal.
	Such a distribution has been demonstrated to be advantageous for improving the robustness against adversarial attacks in previous literature.
	\item The proposed loss simultaneously improves  classification performance and robustness against adversarial attacks. 
	Furthermore, no extra trainable parameters are needed to implement the proposed method.
	\item Extensive experiments with different CNN architectures demonstrate that our method achieves the state-of-the-art performance on detecting adversarial examples.
	It also performs favorably  against the commonly used softmax loss and variants on various visual classification tasks.
\end{itemize}



\section{Related Work}
\label{sec:RW}

\subsection{Loss Function for Classification
}
%
%
Significant efforts have been made to develop effective loss functions other than the softmax loss. 
One of the most widely studied approaches is to explicitly encourage stronger intra-class compactness and larger inter-class separability using the softmax loss.
Sun \etal~\cite{DBLP:conf/nips/SunCWT14} introduce the contrastive loss to a Siamese network for face recognition by simultaneously minimizing the distances between positive image pairs and enlarging the distances between negative image pairs by a pre-defined margin.
Similarly, Schroff \etal~\cite{DBLP:conf/cvpr/SchroffKP15} apply a similar inter-sample distance regularization term on triplets rather than pairs of images. 
One issue with the contrastive loss and triplet loss is the combinatorial explosion in the number of image pairs or triplets especially for large-scale datasets, leading to a significant increase in the required number of training iterations.
The center loss \cite{DBLP:conf/eccv/WenZL016} alleviates the computational loads of pair-wise or triplet-wise by minimizing the Euclidean distance between the features and corresponding class centroids.
However, this formulation introduces inconsistency of distance measurements in the feature space because the class affinity scores are computed based on dot product instead of Euclidean distance.
Similarly, Qian \etal~\cite{softtriple} propose the SoftTriple loss for fine-grained image classification in which multiple centers are learned for each class without sampling triplets.

In general, introducing the classification margin during training is likely to obtain performance gain. 
Liu \etal~\cite{DBLP:conf/icml/LiuWYY16} introduce an angular margin into the softmax loss through the designing of a sophisticated differentiable angular distance function.
Kobayashi~\cite{kobayashiLarge} shows that a classification margin can be more naturally embedded in the formulation of the softmax loss.
However, to leverage such a margin, a strict constraint has to be introduced by optimizing the symmetric Kullback-Leibler divergence between the distribution of the non-ground-truth logits and the uniform distribution.

Several approaches have been developed to improve the numerical stability of the softmax loss. 
The label smoothing \cite{DBLP:journals/corr/SzegedyVISW15} and knowledge distillation \cite{Hinton2015Distilling} methods replace the one-hot ground truth distribution with other distributions that are probabilistically more reasonable.
The approach by Chen \etal~\cite{Chen2017Noisy} focuses on mitigating the early saturation problem of the softmax loss by injecting annealed noise in the softmax function during each training iteration.
The above-mentioned methods aim to improve the softmax loss by external modifications rather than reformulating its fundamental assumption.

\subsection{Robustness against Adversarial Attacks}
Learning robust deep models against adversarial attacks \cite{goodfellow2014explaining} has recently attracted much attention where inputs are specifically combined with imperceptible but worst-case perturbations to cause the model to make incorrect classifications with high confidence.
%
%
Several methods~\cite{goodfellow2014explaining, szegedy2013intriguing, kurakin2016adversarial} add adversarial examples into the training dataset for data augmentation to improve the robustness against adversarial examples.
However, these methods depend on specific attack methods and are vulnerable to iterative attacks~\cite{kurakin2016adversarial}.
Madry \etal~\cite{madry2018towards} propose a more general framework for adversarial training to deal with iterative attack methods.
However, generating adversarial examples by iterative attacks during training is computationally expensive~\cite{madry2018towards}. 

The defensive distillation approach~\cite{papernot2016distillation} aims to train a substitute model with smoother gradients to increase the difficulty of generating adversarial examples.
Nevertheless, it is not effective for the optimization-based attack methods such as~\cite{carlini2017towards}.
Pang~\etal~\cite{pang2018max} propose to model the feature distribution as a Gaussian Mixture in which 
the Gaussian means are preset by carefully determining a sensitive parameter for the vector norm and fixed during training.
Consequently, this method performs well only on simpler datasets (\eg MNIST).
%
In contrast,  our method allows smooth optimization of Gaussian means and performs well for challenging large-scale datasets (\eg ImageNet).

The features of adversarial examples can be modeled by a probability distribution significantly different from that of the clean training samples. 
%
%
%
Numerous methods have been developed to detect adversarial examples.  
However, none of these methods aim to increase the robustness for adversarial attacks while improving classification performance.

{\flushleft \textbf{Auxiliary Trainable Modules.}} 
Metzen \etal~\cite{metzen2017detecting} augment deep models with an auxiliary sub-network to discriminate adversarial examples from the normal ones.
Similarly, the SafetyNet~\cite{lu2017safetynet} trains a support vector machine to classify two types of samples.
However, both methods do not generalize well to unseen types of adversarial attacks.
Besides, they require training additional models.
In contrast, our method is trained only for the classification task  without needing additional models. 

{\flushleft \textbf{Modeling the Input Distribution.}} 
%
The Defense-GAN~\cite{defenseGAN} and PixelDefend~\cite{song2017pixeldefend} methods model the distribution of input image data and distinguish adversarial examples based on the discrepancy of distributions.
However, it is difficult to accurately model the data distribution in the image space due to the noise and large variations in this space. 
In contrast, our method models the distribution in the feature space where the learned representations usually have lower dimensions, richer semantic information and less noise.

{\flushleft \textbf{Modeling the Feature Distribution.}}
In \cite{feinman2017detecting}, Feinman \etal show that adversarial examples can be effectively detected based on the kernel density (K-density) estimation in the feature space of deep models.
Zheng \etal~\cite{zheng2018intrinsic} train a separate set of Gaussian Mixture Models (GMMs) to model the distribution of the extracted deep features for a pre-trained Deep Neural Network (DNN).
The GMMs are then adopted to discriminate adversarial examples based on the likelihood of deep features. 
Both the K-density and GMMs based methods only model the distribution as a post-processing procedure for a pre-trained model.
%
Neither of them optimizes the DNN model by learning a more robust feature distribution against adversarial attacks.

{\flushleft \textbf{Learning the Feature Distribution.}} The Reverse Cross Entropy (RCE) loss~\cite{pang2018rce} is developed to constrain deep features to be within low dimensional manifolds and improve the detection performance by using K-density as a metric to detect adversarial examples in this feature space.
%
However, to obtain these low dimensional manifolds, this method requires the predicted probabilities of classes other than the ground-truth one to be equal, which is a strong assumption and leads to substantial convergence problems for large-scale datasets.
In contrast, the proposed GM loss is scalable to large datasets for both robust detection of adversarial examples and more accurate classification of clean images.

\section{Gaussian Mixture Loss}
\label{method}
In this section, we introduce the GM loss and present
mathematical analysis of the optimal solution. 
We explain why the proposed loss leads to a symmetric feature space which enhances model robustness against adversarial attacks. 
We also describe how to efficiently add a classification margin to the GM loss, which improves the classification performance.

\subsection{GM loss formulation}
\label{sect_formulation}

In contrast to the softmax loss, we assume that the extracted deep feature $x \in \mathbb{R}^D$, follows a Gaussian mixture distribution, 
\begin{equation}
\label{rbfprob1}
p(x) = \sum_{k=1}^{K} \mathcal{N}(x;\mu_k,\Sigma_k) p(k), 
\end{equation}
where $\mu_k$ and $\Sigma_k$ are the mean and covariance of class $k$ in the feature space; and $p(k)$ is the prior probability of class $k$. 
As discussed in Section~\ref{sec:introduction}, the number of Gaussian components is set to the number of object categories $K$ as this work focuses on visual classification.

With this assumption, the conditional probability distribution of the feature $x_i$ of the $i$-th sample, given its class label $z_i \in [1,K]$, can be expressed by
\begin{equation}
\label{rbfprob2}
p(x_i|z_i) = \mathcal{N}(x_i;\mu_{z_i},\Sigma_{z_i}).
\end{equation}
Consequently, the corresponding posterior probability distribution can be expressed by 
\begin{eqnarray}
\begin{aligned}
\label{rbfprob3}
p(z_i|x_i) = \frac{\mathcal{N}(x_i;\mu_{z_i},\Sigma_{z_i})p(z_i)}{\sum_{k=1}^{K}\mathcal{N}(x_i;\mu_k,\Sigma_k) p(k)}.
\end{aligned}
\end{eqnarray}
As such, a \emph{classification loss} $\mathcal{L}_{cls}$ can be computed by the cross-entropy between the posterior probability distribution and one-hot class label,
\begin{eqnarray}
\label{eq_lossi}
\begin{aligned}
\mathcal{L}_{cls} &= -\frac{1}{N}\sum_{i=1}^{N}\sum_{k=1}^{K} \mathbbm{1}(z_i=k) \log {p}(k|x_i) \\
&= -\frac{1}{N}\sum_{i=1}^{N}\log \frac{\mathcal{N}(x_i;\mu_{z_i}, I)}{\sum_{k=1}^{K}\mathcal{N}(x_i;\mu_k, I)},
\end{aligned}
\end{eqnarray}
where the indicator function $\mathbbm{1}()$ equals $1$ if $z_i$ equals $k$; or 0 otherwise.
$N$ is the number of training samples.
For simplicity, we adopt a simple model in which the prior probability $p(k)$ is the constant $1/K$ and the covariance matrix $\Sigma_k$ is the identity matrix $I$.

Optimizing the classification loss alone does not necessarily drive the extracted training features towards the GM distribution.
For example, a feature $x_i$ can be far away from the corresponding class centroid $\mu_{z_i}$ while still being correctly classified as long as it is relatively closer to $\mu_{z_i}$ than to the feature means of the other classes.
To solve this problem, we introduce a \emph{likelihood regularization} term to measure the  extent that a training sample fits the assumed distribution.
%
%
The likelihood under the Gaussian mixture distribution is adopted instead of the likelihood under a single Gaussian component given the class label.
Theoretical analysis for the advantages of this regularization term is presented in Section~\ref{sec_optimal}. 
Formally, the log-likelihood for a feature $x_i$ is given by
\begin{equation}
\label{likelihood2}
\log p(\mathrm{x_i|\mu, I}) = \log\frac{1}{K}\sum_{k=1}^K  \mathcal{N}(x_i;\mu_{k}, I).
\end{equation}
However, this term cannot be directly used because it leads to a trivial optimal solution in which the feature space either scales up towards infinity or collapses to the origin point.
Instead, we maximize its lower bound using Jensen's Inequality, 
%
\begin{equation}
\begin{aligned}
\label{ineq_likelihood}
\log p(\mathrm{x_i|\mu, I}) \geq& \frac{1}{K}\sum_{k=1}^K  \log\mathcal{N}(x_i;\mu_{k}, I) \\
=& -\frac{1}{2K}\sum_{k=1}^K\norm{x_i - \mu_k}^2 - \frac{D}{2}\log 2\pi,
\end{aligned}
\end{equation}
in which $D$ represents the dimension of the feature vector $x_i$. 
The likelihood regularization term can be derived by
\begin{equation}
\label{l_reg}
\mathcal{L}_{lkd} = \frac{1}{K}\sum_{i=1}^N\sum_{k=1}^K\norm{x_i - \mu_{k}}^2,
\end{equation}
where the constant term and constant coefficient in Eq.~\ref{ineq_likelihood} has been ignored. 
Finally, the proposed GM loss $\mathcal{L}_{GM}$ is defined by 
\begin{equation}
\label{l_gm}
\mathcal{L}_{GM} = \mathcal{L}_{cls} + \lambda \mathcal{L}_{lkd},
\end{equation}
where $\lambda$ is a non-negative weighting coefficient. 

By definition, the classification loss $\mathcal{L}_{cls}$ is mainly related to its discriminative capability while the likelihood regularization $\mathcal{L}_{lkd}$ is related to its probabilistic distribution.
Combining these two objectives during the training process helps derive a model with high classification performance and better distribution modeling of features.

\subsection{Optimal Solution for GM loss}
\label{sec_optimal}

\begin{figure}[t]
	\centering 
	\subfigure[MNIST]{\label{fig:a}\includegraphics[width=0.95\linewidth]{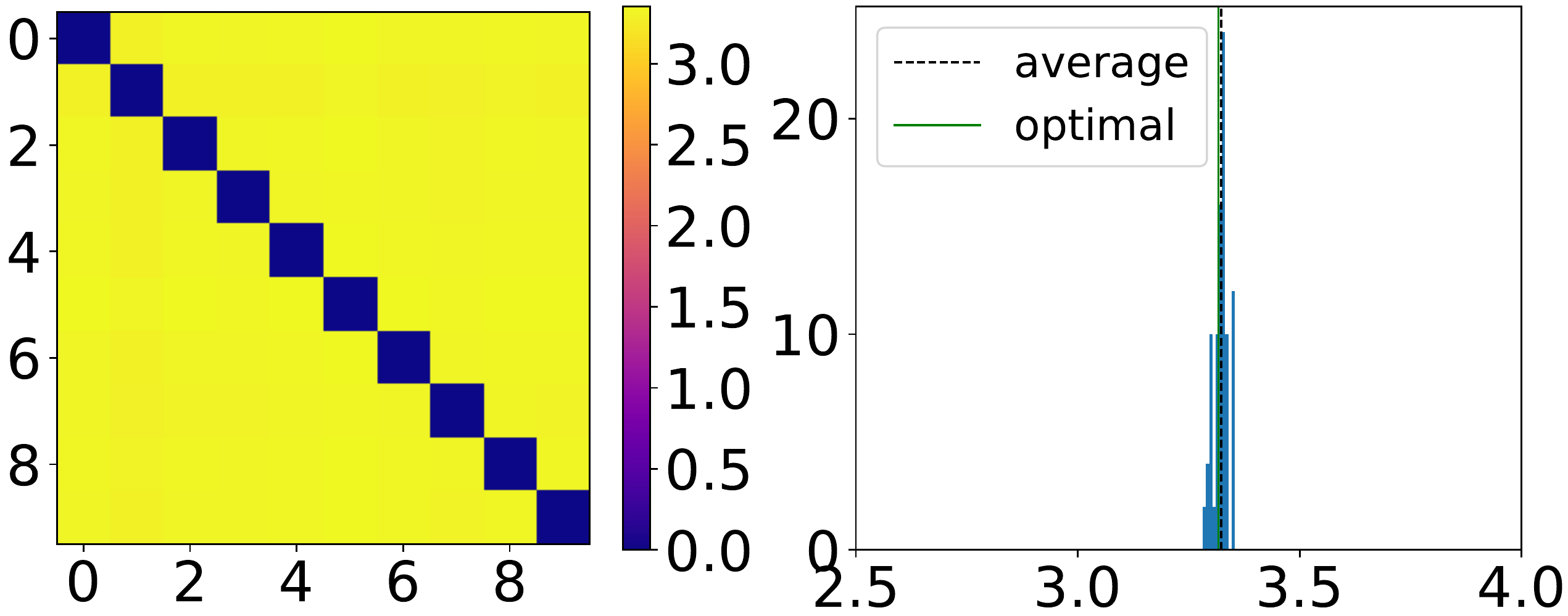}}
	\subfigure[CIFAR10]{\label{fig:b}\includegraphics[width=0.95\linewidth]{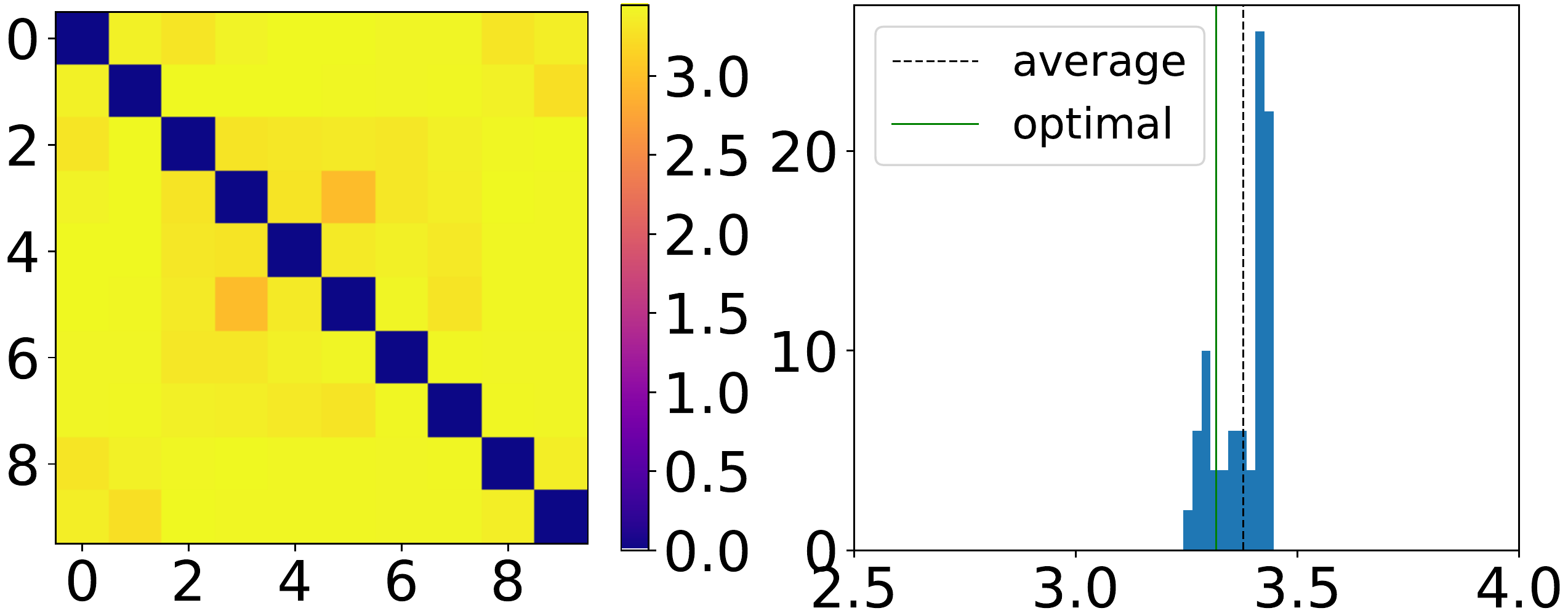}}
	\caption{\textbf{Numerical analysis for ResNet32~\cite{DBLP:conf/cvpr/HeZRS16} model trained with GM loss on the MNIST or CIFAR10 dataset.} Left: the grid on location (m, n) represents $d_{mn}$. Right: the histogram of $d_{mn}, m\neq n$. The black dashed line shows the average value of $d_{mn}$ and the green solid line shows the theoretically optimal value according to Eq.~\ref{eq_minimal}.}
	\label{fig_c2c_distance}
		\vspace{-3mm}
\end{figure}

In this section, we present theoretical analysis of the optimal GM loss for feature learning.
The formulation of the proposed GM loss naturally leads to a symmetric feature space which has been shown to be effective for robust deep models against adversarial attacks~\cite{pang2018max, pang2018rce}.
We also show the optimal distances between Gaussian means of different classes are consistent with experimental results in Fig.~\ref{fig_c2c_distance}.

Suppose there are $N$ training pairs $(x_i, y_i), i=1, 2, \ldots, N$, where $x_i$ is the feature vector of the $i$-th input image and $y_i$ is the corresponding ground-truth label.
%
We further assume that the number of training samples for each class is the same (or we can augment the data points belonging to the classes with fewer samples).
Let the number of classes be $K$, 
the feature vector from the $n$-th class satisfies $x^{(n)} \sim \mathcal{N}(\mu_n, I), n=1, 2, \ldots, K$.
According to the law of large numbers, when the number of samples $N$ is large enough, Eq.~\ref{l_gm} can be expressed as
\begin{eqnarray}
\label{eq_expectation}
\begin{aligned}
\mathcal{L}_{GM} = \frac{1}{K}\sum_{n=1}^{K} \mathbb{E}[&-\log \frac{\mathcal{N}(x^{(n)};\mu_{n},I)}{\sum_{m=1}^{K}\mathcal{N}(x^{(m)};\mu_m, I)} + \\ &\lambda\frac{1}{K}\sum_{m=1}^K\norm{x^{(n)} - \mu_m}^2],
\end{aligned}
\end{eqnarray}
where $\mathbb{E}(\cdot)$ is the expected value. 
We analyze the two terms in Eq.~\ref{eq_expectation} separately.
Since $x^{(n)} \sim \mathcal{N}(\mu_n, I)$, the second term in Eq.~\ref{eq_expectation}, denoted by $Loss_{reg}$, can be computed by
\begin{eqnarray}
\label{eq_expec_reg}
\begin{aligned}
Loss_{reg} &= \frac{\lambda}{K^2}\sum_{n=1}^{K}\sum_{m=1}^{K}\mathbb{E}[(x^{(n)}-\mu_m)^\top (x^{(n)}-\mu_m)] \\
&= \frac{\lambda}{K^2}\sum_{n=1}^{K}\sum_{m\neq n}^{K}d_{mn}^2 + D,
\end{aligned}
\end{eqnarray}
in which $d_{mn} = \norm{\mu_m-\mu_n}$.
As such, $Loss_{reg}$ is directly related to the distances between different Gaussian means.

Denote the first term in Eq.~\ref{eq_expectation} by $Loss_{cls}$.
Let $f_n(x)=\mathcal{N}(x;\mu_n, I)$ and $g_n(x)=\frac{1}{K}\sum_{m=1}^{K}\mathcal{N}(x;\mu_m, I)$, 
we have
\begin{eqnarray}
\label{eq_expec_cls}
\begin{aligned}
Loss_{cls} &= \frac{1}{K}\sum_{n=1}^{K} \mathbb{E}[-\log \frac{\mathcal{N}(x^{(n)};\mu_{n},I)}{\sum_{m=1}^{K}\mathcal{N}(x^{(m)};\mu_m, I)}] \\
&= \frac{1}{K}\sum_{n=1}^{K}\mathbb{E}[-\log \frac{f_n(x^{(n)})}{Kg_n(x^{(n)})}] \\
&= \log{K} - \frac{1}{K}\sum_{n=1}^{K}\dkl{f_n(x)}{g_n(x)},
\end{aligned}
\end{eqnarray}
where $D_{KL}$ denotes the Kullback-Leibler divergence.
As shown in~\cite{durrieu2012lower}, the KL divergence between a Gaussian component and a Gaussian mixture can be computed by the variational approximation, 
\begin{equation}
\label{varapprox}
D_{var}(f_n\Vert g_n) = -\log (\frac{1}{K}\sum_{m=1}^{K}e^{-\dkl{\mathcal{N}(x;\mu_n, I)}{\mathcal{N}(x;\mu_m, I)}}). 
\end{equation}
%
%
%
As such, an approximation of $Loss_{cls}$ can be obtained by 
\begin{eqnarray}
\label{eq_expec_cls_approx}
\begin{aligned}
Loss_{cls} &\approx \log{K} - \frac{1}{K}\sum_{n=1}^{K}D_{var}(f_n(x)\Vert g_n(x)) \\
&=\frac{1}{K}\sum_{n=1}^{K}\log( \sum_{m=1}^K e^{-\frac{1}{2}d_{mn}^2}). 
\end{aligned}
\end{eqnarray}
Based on Eq.~\ref{eq_expec_reg} and Eq.~\ref{eq_expec_cls_approx}, the GM loss in Eq.~\ref{eq_expectation} can be represented as a function of $d_{mn}$ by
%
%
\begin{eqnarray}
\begin{aligned}
\label{eq_loss_wrt_d}
\mathcal{L}_{GM} \approx \frac{1}{K}\sum_{n=1}^{K}&[\log(\sum_{m=1}^K e^{-\frac{1}{2}d_{mn}^2}) \\ 
& + \lambda(\frac{1}{K}\sum_{m\neq n}^{K}d_{mn}^2 + D)].
\end{aligned}
\end{eqnarray}
The optimal values of  of $d_{mn}$ for minimizing $\mathcal{L}_{GM}$ can be analytically solved as shown below.
%
Let $d_n^2 = \frac{1}{K-1}\sum_{m\neq n}^{K}d_{mn}^2$.
Considering that $f(x) = e^{-x/2}$ is a convex function,
by applying the Jensen's Inequality, we have 
\begin{equation}
\mathcal{L}_{GM} \geq 1/K\sum_{n=1}^{K}[\log (1+(K-1)e^{-d_n^2/2}) + \lambda (d_n^2(K-1)/K +D)]. 
\label{jensen}
\end{equation}
The equality holds if and only if $\forall n, d_{mn} = d_n (m\neq n)$.

Let $d^2=1/K\sum_{n=1}^K d_n^2$.
The function $f(x) = \log (1 + (K-1)e^{-x/2})$ is a convex function.
Based on inequality~\ref{jensen}, by Jensen's inequality, we have
\begin{eqnarray}
\begin{aligned}[b]
\label{eq_minimal}
\mathcal{L}_{GM} \geq &\log (1 + (K-1)e^{-d^2/2}) + \lambda (\frac{K-1}{K}d^2 + D). 
\end{aligned}
\end{eqnarray}
The equality holds if and only if $\forall n, d_n=d$.

The minimal value for $\mathcal{L}_{GM}$, which is given in Eq.~\ref{eq_minimal}, is reached if and only if $\forall m\neq n, d_{mn}=d$.
In other words, training deep models with the proposed GM loss naturally learns a symmetric feature space in which the distances between different Gaussian means are all equal.
Empirical results in Fig.~\ref{fig_c2c_distance} demonstrate this property where the distances between different Gaussian means are approximately equal.
We note that the actual values of $d_{mn}$ are not strictly equal to each other.
As the number of training samples is finite and the model capacity is limited, it is difficult to perfectly transform the data distribution in the input space to the assumed distribution in the feature space.
Nevertheless, Fig.~\ref{fig_c2c_distance} shows reasonable consistency between the theoretical analysis and empirical results.

In~\cite{pang2018max}, it has been shown that such a symmetric feature space intrinsically enhance model robustness against adversarial attacks because it maximizes the minimal distance between different classes.
It is worth noticing that a symmetric feature space is naturally learned under the supervision of GM loss instead of introducing explicit constraints~\cite{pang2018rce} or constructing hand-crafted fixed Gaussian means~\cite{pang2018max} throughout the training process. 

%

\subsection{Large-Margin GM Loss}
In addition to the model robustness against adversarial attacks, the classification performance can be improved using the GM loss with some modification. 
It has been widely recognized in statistical learning that large classification margin on the training set usually helps generalization, which has been applied to deep models \cite{DBLP:conf/aaai/SunCWLL16,DBLP:conf/icml/LiuWYY16}.
Denote the contribution of $x_i$ to the classification loss to be $\mathcal{L}_{cls,i}$.
Based on Eq.~\ref{eq_lossi}, $\mathcal{L}_{cls,i}$ is described by
\begin{equation}
\label{l_cls_i}
\mathcal{L}_{cls,i} = -\log \frac{p(z_i)|\Sigma_{z_i}|^{-\frac{1}{2}}e^{-s_{z_i}}}{\sum_{k}p(k)|\Sigma_k|^{-\frac{1}{2}}e^{-s_k}},
\end{equation}
where 
\begin{equation}
\label{l_cls_d}
s_k = (x_i - \mu_k)^\top\Sigma_{k}^{-1}(x_i - \mu_k) / 2. 
\end{equation}
Since the squared Mahalanobis distance $s_k$ is by definition non-negative, a classification margin $m$ ($m \geq 0$) can be easily introduced to achieve the large-margin GM loss as in 
\begin{equation}
\label{l_cls_m}
\mathcal{L}_{cls,i}^{m} = - \log \frac{p(z_i)|\Sigma_{z_i}|^{-\frac{1}{2}}e^{-s_{z_i}-m}}{\sum_{k}p(k)|\Sigma_k|^{-\frac{1}{2}}e^{-s_k-\mathbbm{1}(k=z_i)m}}. 
\end{equation}
Clearly, adding the classification margin to the GM loss is more straightforward than to the softmax loss \cite{DBLP:conf/icml/LiuWYY16}.   
%


Since we consider the case in which $p(k)$ is constant and $\Sigma_k$ is the identity matrix, then $x_i$ is classified to the class $z_i$ if and only if Eq.~\ref{explain_margin} holds.
This indicates that $x_i$ should be closer to the feature mean of class $z_i$ than to the one of the other classes by at least a margin $m$, \ie
\begin{equation}
\label{explain_margin}
e^{-s_{z_i} - m} > e^{-s_k} \Longleftrightarrow s_k - s_{z_i} > m \quad ,\forall k \ne z_i
\end{equation}
Furthermore, we adopt an adaptive scheme by setting the value of $m$ to be proportional to each sample's distance to its corresponding Gaussian mean, \ie\  $m = \alpha s_{z_i}$, in which $\alpha$ is a non-negative parameter controlling the size of the expected margin between two classes on the training set.

\begin{figure}[t]
	\centering
	\includegraphics[width=0.99\linewidth]{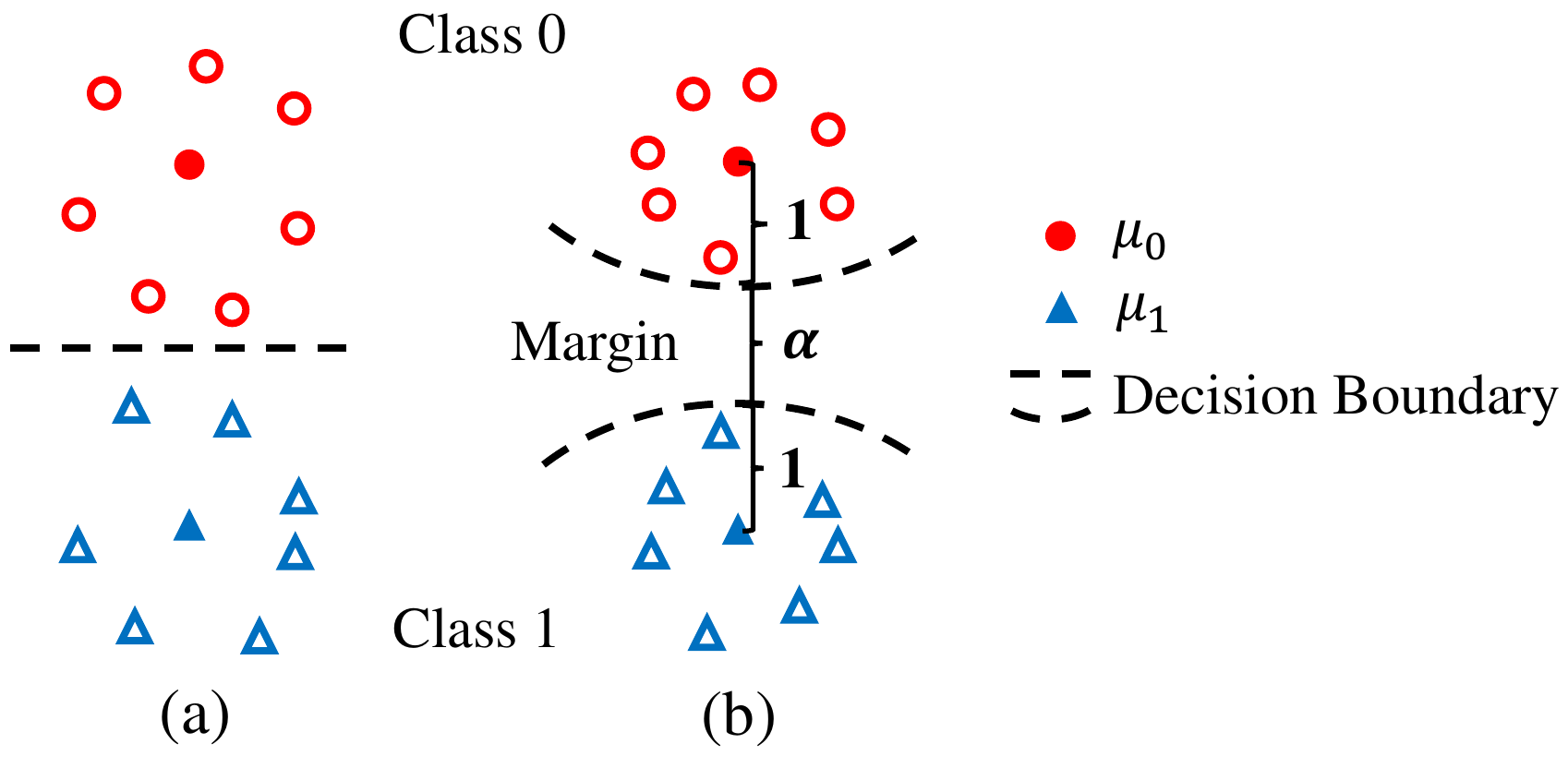}
	\caption{\textbf{Geometric interpretation of the relationship between \bm{$\alpha$} and the margin size in the training feature space.} (a) GM loss with $\alpha=0$; (b) GM loss with $\alpha>0$.} 
	\label{fig_geometry}
\end{figure}

It should be emphasized that this formulation cannot be directly applied to the softmax loss since an inner-product affinity score can be positive or negative, whereas a sensible margin generally has to be non-negative.
Fig.~\ref{fig_geometry} shows a schematic interpretation of $\alpha$; and Fig.~\ref{fig01}(d) and (e) illustrate how the feature space changes when increasing $\alpha$ from $0$ to $1$.
The margin does not change the symmetry property described in Section~\ref{sec_optimal} because it reduces the distance between features and the their Gaussian mean instead of changing the distance between different Gaussian means.

\section{Experimental Results}
\label{sec_exp}

We evaluate the proposed method using two sets of experiments.
First, we conduct image classification and face verification experiments to demonstrate the effectiveness of the large-margin GM loss (GM loss for short).
We present the mean and standard deviation of three runs except for the large-scale dataset, \ie ImageNet.
Next, we demonstrate the effectiveness of detecting adversarial examples using kernel density estimation.
All experiments are carried out using the \emph{Tensorflow}~\cite{tensorflow} on NVIDIA TitanX GPUs. 
The source code and trained models will be made available to the public.

\subsection{Image Classification}
\label{sect_exp_visual}

{\flushleft \textbf{MNIST.}} We first evaluate the softmax loss, center loss (with the softmax loss) \cite{{DBLP:conf/eccv/WenZL016}}, large-margin softmax loss (L-Softmax loss for short) \cite{DBLP:conf/cvpr/SchroffKP15} and GM loss by visualizing the learned 2D feature spaces for the MNIST Handwritten Digit dataset \cite{mnist}.
We use a network with 6 convolution layers and a fully connected layer with a two-dimensional output.
The feature embeddings on the training set with different loss functions are illustrated in Fig.~\ref{fig01}. 
Different from the softmax loss and its variants, the features generated using the GM loss generally follow the GM distribution, which is consistent with the assumption of this work.
The heatmap of the kernel density in the feature space is shown in  Fig.~\ref{fig01}(f).
In addition, larger margin sizes can be observed among different classes
with an increasing $\alpha$ as shown in 
Fig.~\ref{fig01}(d)-(e).

For quantitative evaluations, we increase the output dimension of the fully connected layer from 2 to 100 and add a ReLU activation.
For fair comparisons, we train the same network with different loss functions using the same training parameters, including the learning rate and weight decay.
The classification error rate on the test set is presented in Table~\ref{tb_mnist}.

\begin{table}[t]
	\centering
	\caption{\textbf{Classification error rates (\%) on the MNIST test set.} The models are trained using a 6-layer CNN with different loss functions.}
	\label{tb_mnist}
	\begin{tabular}{ccc}
		\hline
		Loss Functions & 2-D (\%) & 100-D (\%) \\
		\hline
		Center Loss\cite{{DBLP:conf/eccv/WenZL016}}  & 1.45 $\pm$ 0.01 & 0.47 $\pm$ 0.01 \\
		L-Softmax Loss\cite{DBLP:conf/icml/LiuWYY16} & 1.30 $\pm$ 0.02 & 0.43 $\pm$ 0.01 \\
		Softmax Loss& 1.82 $\pm$ 0.01 & 0.68 $\pm$ 0.01 \\
		\hline
		GM Loss($\alpha=0$)) & 1.44 $\pm$ 0.01 & 0.49 $\pm$ 0.01\\
		GM Loss($\alpha=0.3$)  & 1.32 $\pm$ 0.01 & 0.42 $\pm$ 0.02\\
		GM Loss($\alpha=1.0$) & \textbf{1.17 $\pm$ 0.01} & \textbf{0.39 $\pm$ 0.01} \\
		\hline
	\end{tabular}
\end{table}

{\flushleft \textbf{CIFAR.}}  The CIFAR-10 dataset \cite{krizhevsky2009learning} consists of $32\times 32$ pixel color images from 10 classes, with 50,000 training images and 10,000 testing images.
The CIFAR-100 dataset \cite{krizhevsky2009learning} has 100 classes containing 600 images each. 
There are 500 training images and 100 testing images per class. 
%
We use the typical data augmentation method including mirroring and $32\times 32$ random cropping after 4-pixel reflection paddings on each side \cite{DBLP:conf/cvpr/HeZRS16, DBLP:conf/icml/LiuWYY16}.

For CIFAR-10, We train the ResNet \cite{DBLP:conf/cvpr/HeZRS16} of 20, 56 and 110 layers with different loss functions. 
The networks are trained with a batch size of 128 for 300 epochs, and the learning rate is set to 0.1 and then divided by 10 at the $150^{th}$ epoch and the $225^{th}$ epoch respectively. 
We use a weight decay of $5\times 10^{-4}$ and the Nesterov optimization algorithm \cite{sutskever2013importance} with a momentum of 0.9.
The network weights are initialized using the method introduced in \cite{he2015delving}.
The recognition error rate is shown in Table~\ref{tb_cifar10}.
The results in the first row are reported in the original RestNet paper \cite{DBLP:conf/cvpr/HeZRS16}.
For the center loss and the large-margin softmax loss, we train the models by ourselves since the ResNet was not used on the CIFAR-10 dataset in  \cite{DBLP:conf/eccv/WenZL016} and \cite{DBLP:conf/icml/LiuWYY16}. 
The model with the GM loss performs favorably against the softmax loss and its two variants for different ResNet models of different depths.

\begin{table}[t]
	\caption{\textbf{Classification error rates (\%) on the CIFAR-10 dataset.} The models are trained using the ResNet architectures with different loss functions.}
	\label{tb_cifar10}
	\small
	\centering
	\begin{tabular}{cccc}
		\hline
		Loss Functions & ResNet-20 & ResNet-56 & ResNet-110 \\
		\hline
		\hline
		Softmax \cite{DBLP:conf/cvpr/HeZRS16} & 8.75 $\pm$ 0.04 & 6.97 $\pm$ 0.05 & 6.43 $\pm$ 0.04 \\
		Center \cite{DBLP:conf/eccv/WenZL016} & 7.77 $\pm$ 0.05 & 5.94 $\pm$ 0.02 & 5.32 $\pm$ 0.03 \\
		L-Softmax \cite{DBLP:conf/icml/LiuWYY16} & 7.73 $\pm$ 0.03 & 6.05 $\pm$ 0.04 & 5.79 $\pm$ 0.02 \\
		\hline
		GM($\alpha=0.3$) & \textbf{7.21 $\pm$ 0.04} & \textbf{5.61 $\pm$ 0.02} & \textbf{4.96 $\pm$ 0.03} \\
		\hline
	\end{tabular}  
\end{table}

For the CIFAR-100 dataset, we use the same CNN architecture used by the large-margin softmax loss \cite{DBLP:conf/icml/LiuWYY16}, which has the same network architecture as the VGG-net \cite{DBLP:journals/corr/SimonyanZ14a} with 13 convolutional layers and one fully connected layer. 
Bach normalization \cite{DBLP:journals/corr/IoffeS15} is used after each convolutional layer and no dropout is used.
To achieve better recognition performances, we replace the fully connected layer in this network with the global average pooling operator \cite{DBLP:journals/corr/LinCY13}.
We present the recognition results with or without the data augmentation in Table \ref{tb_cifar100}, denoted by C100+ and C100, respectively.
%
We first note that the model with the proposed GM loss performs favorably against the model with the softmax based losses on both C100+ and C100 datasets.
For the augmented C100+ dataset, increasing the margin parameter $\alpha$ consistently helps achieve better recognition performance. 
However, this is not true for the C100 dataset without data augmentation.
This can be attributed to that the number of training samples for each object class is as low as 500 on C100.
%
As such, the margin size $\alpha$ and model generalization capability is less correlated.

\begin{table}[t]
	\caption{\textbf{Classification error rates (\%) on the CIFAR-100 dataset.} The models are trained using a VGG-like CNN containing 13 convolutional layers with different loss functions.}
	\label{tb_cifar100}
	\centering
	\begin{tabular}{ccc}
		\hline
		Loss Functions & C100 & C100+ \\
		\hline
		\hline
		Center \cite{{DBLP:conf/eccv/WenZL016}} & 24.85 $\pm$ 0.06 & 21.05 $\pm$ 0.03\\
		L-Softmax \cite{DBLP:conf/icml/LiuWYY16} & 24.83 $\pm$ 0.05 & 20.98 $\pm$ 0.04 \\
		Softmax & 25.61 $\pm$ 0.07 & 21.60 $\pm$ 0.04 \\
		\hline
		GM($\alpha=0.1$)  & 23.74 $\pm$ 0.08 & 20.94 $\pm$ 0.03\\
		GM($\alpha=0.2$)  & \textbf{23.04 $\pm$ 0.08} & 20.85 $\pm$ 0.04 \\
		GM($\alpha=0.3$)  & 23.16 $\pm$ 0.05 & \textbf{20.76 $\pm$ 0.03} \\
		\hline 
	\end{tabular}
\end{table}

{\flushleft \textbf{ImageNet.}} We analyze the performance of various CNN architectures combined with different loss functions
on large-scale image classification using the ImageNet (ILSVRC 2012)  dataset \cite{deng2009imagenet}.
For fair comparisons,  we train all the models for 100 epochs on 4 Titan GPUs with a mini-batch size of 64 for each GPU.
The learning rate is initialized as 0.1 and divided by 10 at the $30^{th}$, $60^{th}$ and $90^{th}$ epochs, respectively.
We use a weight decay of 0.0002 and a momentum of 0.9, and no dropout \cite{dropout} is used.
We evaluate the performances for a single $224\times224$ center crop for each image on the ILSVRC 2012 validation set. 
Table \ref{tb_imagenet} shows that the proposed method facilitates achieving better classification results on the large-scale dataset when combined with different CNN architectures.


\begin{table}[t] 
	\caption{\textbf{Classification error rates (\%) on ILSVRC 2012 validation set.} For GM loss, we set the margin parameter $\alpha$=0.01 and regularization parameter $\lambda$=0.1.}
	\label{tb_imagenet}
	\small
	\centering  
	\begin{tabular}{ccccc}  
		\hline  
		\multirow{2}{*}{CNN Architecture}&
		\multicolumn{2}{c}{Softmax loss}&\multicolumn{2}{c}{GM loss}\cr\cline{2-5}  
		&top-1 &top-5 &top-1 &top-5 \cr 	 
		\hline      
		ResNet-50~\cite{resnet} &24.7 &7.8 &\textbf{23.28} &\textbf{6.81}\cr
		ResNet-101~\cite{resnet}&23.6& 7.1& \textbf{22.51}& \textbf{6.22}\cr
		ResNeXt-50~\cite{xie2017aggregated}&22.2 &- &\textbf{21.43} &\textbf{5.72}\cr
		SE-ResNet-50~\cite{senet}&23.29& 6.62 &\textbf{22.37} &\textbf{5.85}\cr
		\hline  
	\end{tabular} 
\end{table} 

\subsection{Face Verification}

We conduct face verification experiments on the Labeled Face in the Wild (LFW) dataset \cite{huang2007labeled}, which contains 13,233 face images of $5749$ different identities with large variations in pose, expression and illumination.
The provided 6,000 pairs are used for face verification test.
We follow the \emph{unrestricted, labeled outside data} protocol 
and use only the CASIA-WebFace dataset \cite{DBLP:journals/corr/YiLLL14a} for training. %
The CASIA-WebFace dataset consists of 494,414 face images from 10,575 subjects.
The training and testing images are aligned using MTCNN \cite{zhang2016joint} and resized to $128\times 128$ pixel.
%
A simple data augmentation scheme is adopted, which includes horizontal mirroring and $120\times 120$ random crop from the aligned $128\times 128$ pixel face images.

\begin{table}[t]
	\caption{\textbf{Face verification performances on the LFW dataset using a single model}. 
	The six models at the bottom are trained on our training scheme while those on top are reported results.}
	\label{tb_face}
	\centering
	\begin{tabular}{ccc}
		\hline
		Method & Training Data & Accuracy (\%) \\
		\hline
		\hline
		FaceNet \cite{schroff2015facenet} & 200M & \textbf{99.65} \\
		Deepid2+ \cite{sun2015deeply} & 0.3M & 98.70 \\ 
		\hline
		Softmax & 0.49M & 98.56 $\pm$ 0.03\\
		L-Softmax \cite{DBLP:conf/icml/LiuWYY16} & 0.49M & 98.92 $\pm$ 0.03\\
		Center \cite{DBLP:conf/eccv/WenZL016} & 0.49M & 99.05 $\pm$ 0.02\\
		GM ($\alpha=0.001$) & 0.49M & 99.12 $\pm$ 0.03\\
		GM ($\alpha=0.005$) & 0.49M & 99.24 $\pm$ 0.02\\
		GM ($\alpha=0.01$) & 0.49M & \textbf{99.41 $\pm$ 0.03} \\
		\hline
	\end{tabular}
\end{table}

We train a face recognition model based on 
the ResNet \cite{DBLP:conf/cvpr/HeZRS16}  
with 27 convolutional layers. 
The PReLU activations \cite{he2015delving} are used after each convolutional layer and no batch normalization or dropout is used.
The model is trained with a batch size of 256 for 20 epochs.
The learning rate is initially set to 0.1 and divided by 10 at the 10th, 14th and 16th epochs.
In the training process, we use the stochastic gradient descent (SGD) method with a momentum of 0.9 and a weight decay of $5\times 10^{-4}$. 
In test phase, we perform principal component analysis to project features from  512 to 64 dimensions and then compute the $L_2$ distance for verification.
For fair comparisons, the verification performance is evaluated on a single model and model ensemble is not used. 
In Table \ref{tb_face}, we list the reported accuracy for the Deepid2+ (contrastive loss) \cite{sun2015deeply} and FaceNet (triplet loss) \cite{schroff2015facenet} methods. 
The FaceNet achieves the highest accuracy of 99.65\% by using a very large training set of 200M images.
%
The reported accuracy of the center loss~\cite{DBLP:conf/eccv/WenZL016} method is 99.28\% 
by using both the CASIA-Webface and  Celebrity+ \cite{Liu2014Deep} datasets for training (with 0.7M training images in total).
When using the CASIA-Webface training dataset only, as shown in Table~\ref{tb_face}, the model with the GM loss performs favorably against the schemes with other loss functions.


%
%

\begin{figure}[t]
	\centering
	\subfigure[Accuracy vs. $\alpha$]{
		\label{fig_param:a}
		\includegraphics[width=0.47\linewidth]{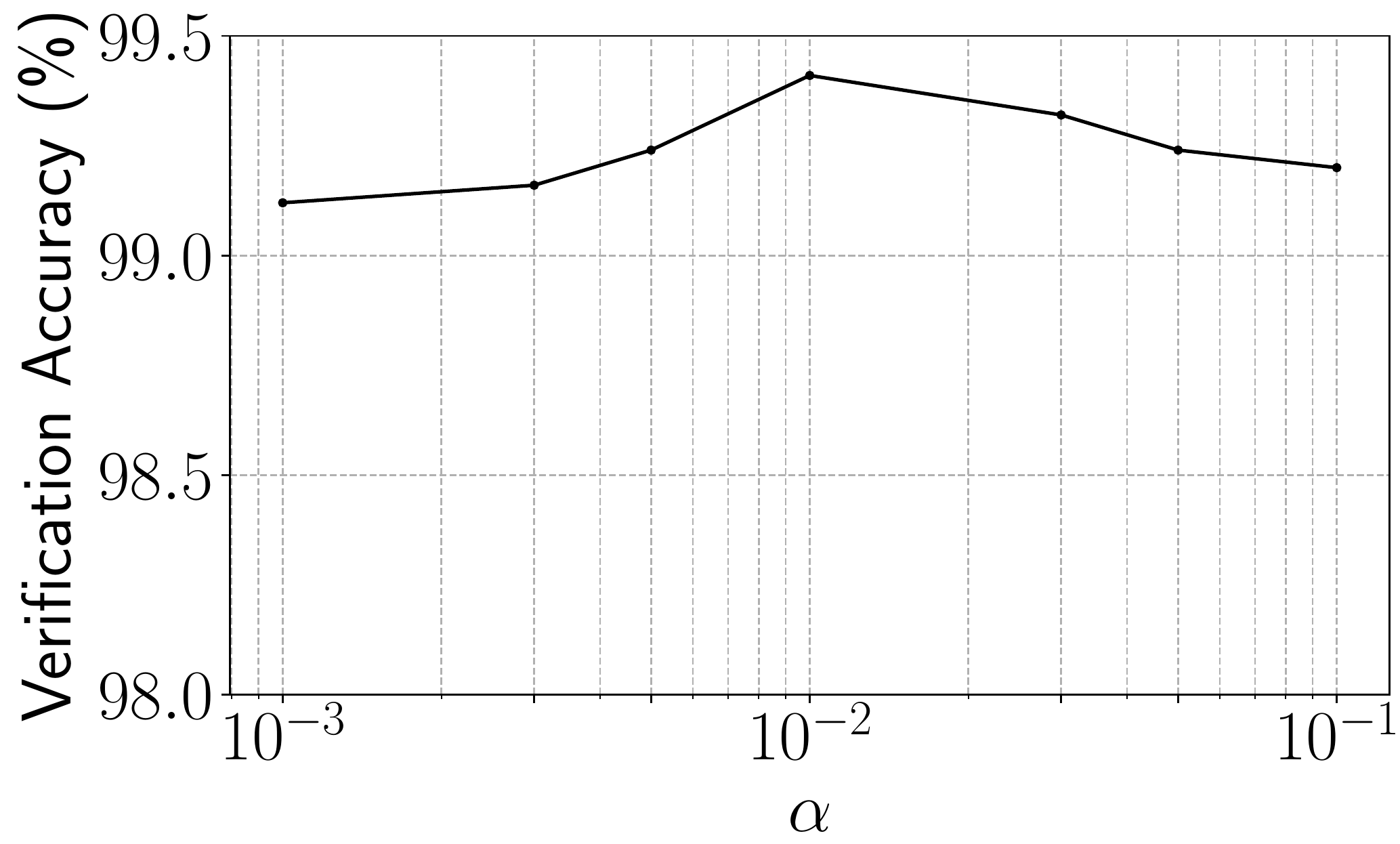}
		}
	\subfigure[Accuracy vs. $\lambda$]{
		\label{fig_param:b}
		\includegraphics[width=0.47\linewidth]{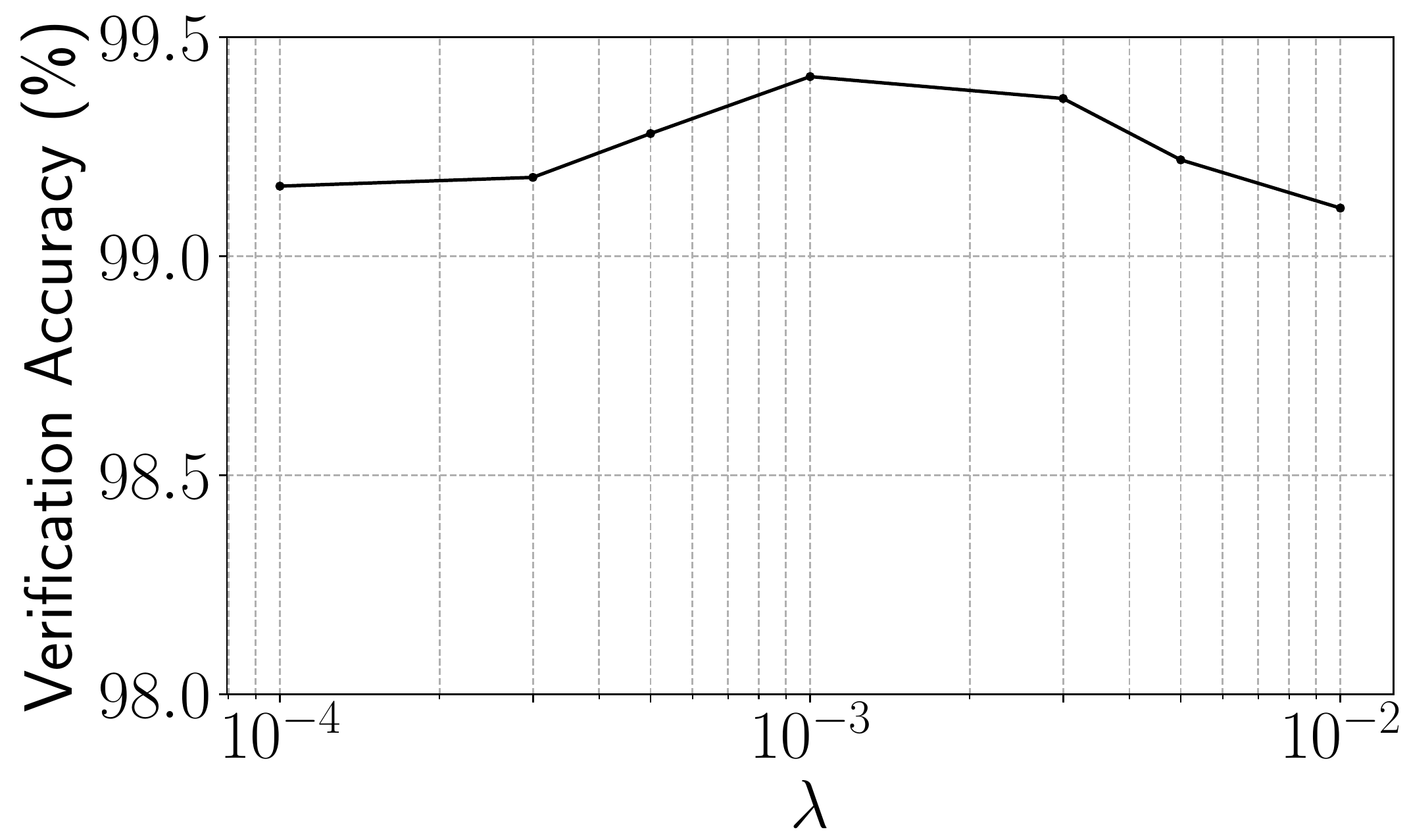}
		}
	\caption{\textbf{Face verification accuracy on the LFW dataset.} (a) models with different $\alpha$ and fixed $\lambda$ = 0.001. (b) models with different $\lambda$ and fixed $\alpha$ = 0.01.} 
	\label{fig_param}
\end{figure}

\begin{table}[t]
	\caption{\textbf{Classification accuracy (\%) on the CIFAR-100 dataset with data augmentation layers.} 
	We analyze the sensitivity of the proposed method with respect to $\lambda$ and $\alpha$.}
	\label{tb_cifar100_ablation}
	\centering
	\begin{tabular}{c|c|c|c|c|c|c}
		\hline
		\diagbox{$\lambda$}{$\alpha$} & 0 & 0.1 & 0.2 & 0.3 & 0.4 & 0.5 \\
		\hline
		0.01&78.40&78.63&78.55&79.08&78.74&78.65\\
		\hline
		0.1&78.62&78.80&78.86&\textbf{79.24}&78.95&78.81\\
		\hline
	\end{tabular}
\end{table}

\subsection{Sensitivity Analysis}
In this section, we conduct experiments on the CIFAR-100 and LFW datasets to study the effect of the margin parameter $\alpha$ and  likelihood regularization parameter $\lambda$ on classification and verification, and discuss how these hyperparameters should be chosen for different datasets.
The experimental results are presetned in Table~\ref{tb_cifar100_ablation} and Fig.~\ref{fig_param} in which the x-axis is displayed in the log scale.

In Fig.~\ref{fig_param:a}, we set $\lambda$ = 0.001 and vary $\alpha$ from $10^{-3}$ to $10^{-1}$. 
For $\alpha$ = 0, \ie without margin term, the verification accuracy is 98.80\%, which shows introducing the margin in the GM loss plays an important role in achieving better  verification performance.
In addition, the model is not sensitive to  $\alpha$ within a wide range.
In Fig.~\ref{fig_param:b}, we set $\alpha$ = 0.01 and vary $\lambda$ from $10^{-4}$ to $10^{-2}$.
For $\lambda$ = 0, the verification accuracy is 98.88\%.
The performance is roughly stable within this range.
Sensitivity analysis is also carried out on the CIFAR-100 classification experiments (see Table~\ref{tb_cifar100_ablation}).

We set $\lambda$ = 0.1 for all the image classification experiments and $\lambda$ = 0.001 for the face verification experiments.
As the face training set has clearly a larger number of classes (over 10,000) than the image classification training set,
the classification loss $\mathcal{L}_{cls}$ for face recognition should have a relatively larger weight, meaning relatively smaller $\lambda$. 
Otherwise, it will be difficult to separate the features of different classes during training.
In general, the likelihood regularization parameter $\lambda$ should be set a small value to prevent the features from collapsing to the origin when the training process starts.
In this setting,  the likelihood regularization starts to play a major role when the training accuracy is approaching saturation.
For the margin parameter $\alpha$, a larger value may lead to a more difficult optimization objective.
Intuitively, $\alpha$ should be smaller when the number of classes gets larger.
In our experiments, we empirically set $\alpha$ to 1.0, 0.3, 0.01 and 0.001 for the MNIST, CIFAR-10, ImageNet and LFW face verification datasets, respectively, in ascending order of the number of classes in the dataset.


\subsection{Adversarial Example Detection}

As discussed in Section~\ref{sec_optimal}, training a network with the proposed GM loss theoretically leads to a symmetric feature space which is more robust against adversarial attacks.
%
As such, we use the kernel density estimation to perform adversarial example detection based on the discrepancy between feature distributions of adversarial examples and normal samples.
We evaluate the performance of the proposed method for adversarial example detection on benchmark datasets, including MNIST~\cite{mnist}, CIFAR-10~\cite{cifar} and  ImageNet~\cite{imagenet} for various adversarial attack methods.
The detection performance is measured by the area under curve (AUC) of the ROC curve for discriminating the normal test samples from the adversarial examples generated on the test set.
We consider the following three types of widely studied threat models:
\begin{enumerate}
	\item{Semi white-box Attack:} Attackers know all the details of the DNN model but are not aware of the defense strategy.
	\item{White-box Attack:} Attackers know all the details of the DNN model and the defense strategy. The adversarial examples are generated to attack the DNN model and defense strategy simultaneously.
	\item{Black-box Attack:} Attackers know all the details of the defense strategy, but do not know the DNN architecture and have no access to the model weights. Adversarial examples are generated based on the gradients of a substitute model.
\end{enumerate}


{\flushleft \textbf{Network Architectures.}} 
For experiments on the MNIST and CIFAR datasets, the ResNet-32~\cite{resnet} model is employed as the target model for adversarial attack.
For experiments on the ImageNet dataset, the ResNet-50 model 
is used for experiments.
Each DNN model is trained with the softmax loss and GM loss, respectively.

{\flushleft \textbf{Adversarial Attacks.}}  We focus on targeted adversarial attacks of which the target classes are chosen randomly from the classes other than the ground-truth class.
For the FGSM~\cite{goodfellow2014explaining}, the $l_\infty$ constraint $\epsilon = 0.1$ for for images with pixel values in range $[-0.5, 0.5]$.
For the gradient-based iterative methods, including BIM and ILCM~\cite{bim}, the number of iterations is set to 20 and the step size $\alpha$, is $0.01$.
For the C\&W attack, the learning rate is set to $0.001$, the binary search steps are five and the Adam~\cite{adam} optmization algorithm is used. 
The confidence threshold $\kappa$ is set to $0$ or $10$, in which the latter case is referred to as C\&W-hc, representing C\&W attack with high confidence.

\begin{figure}[htb]
	\centering
	
	\subfigure{
		\begin{minipage}[t]{0.48\linewidth}
			\centering
			\includegraphics[width=1\linewidth]{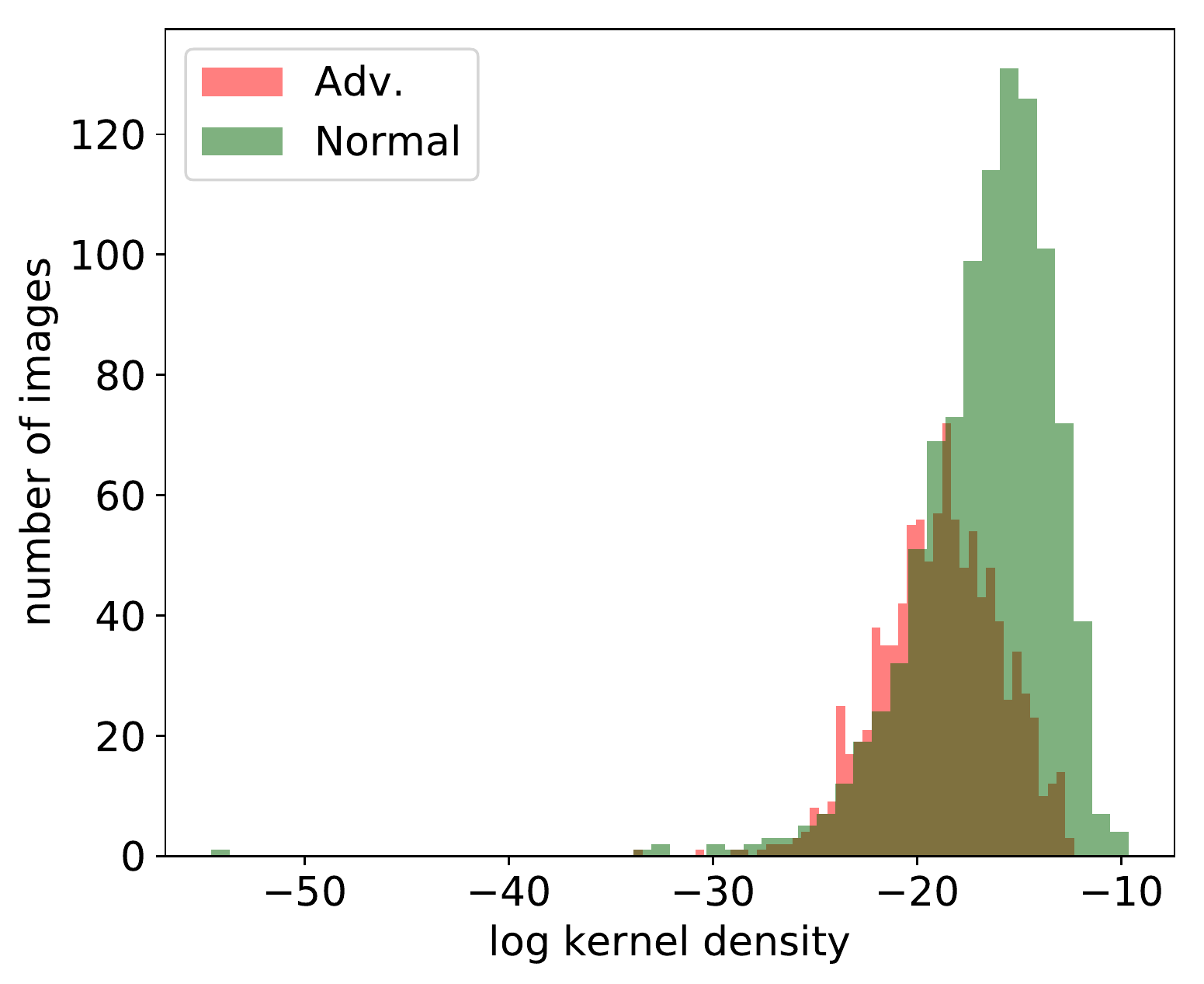}
	\end{minipage}}
	\subfigure{
		\begin{minipage}[t]{0.48\linewidth}
			\centering
			\includegraphics[width=1\linewidth]{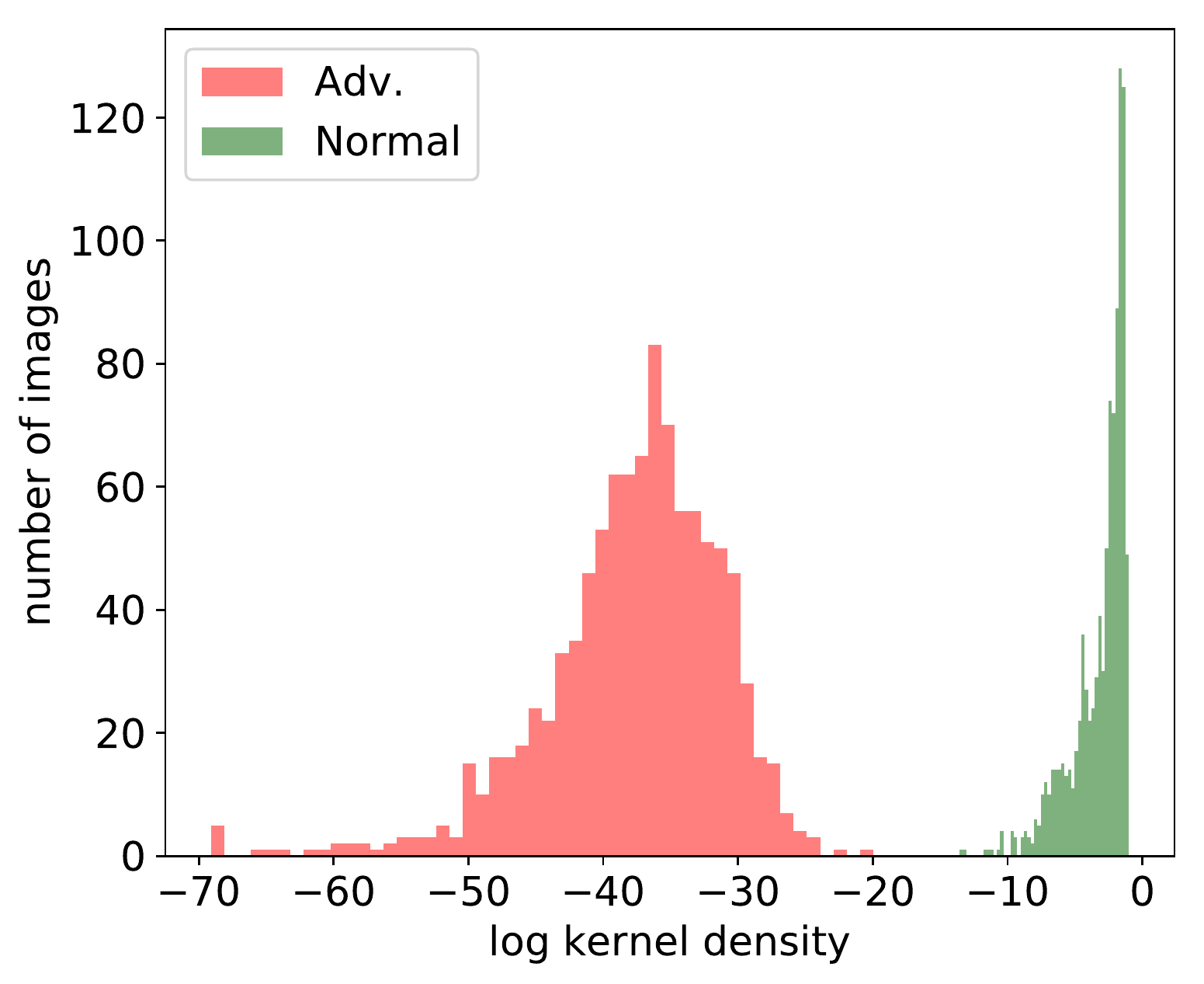}
	\end{minipage}}
	
	\subfigure{
		\begin{minipage}[t]{0.48\linewidth}
			\centering
			\includegraphics[width=1\linewidth]{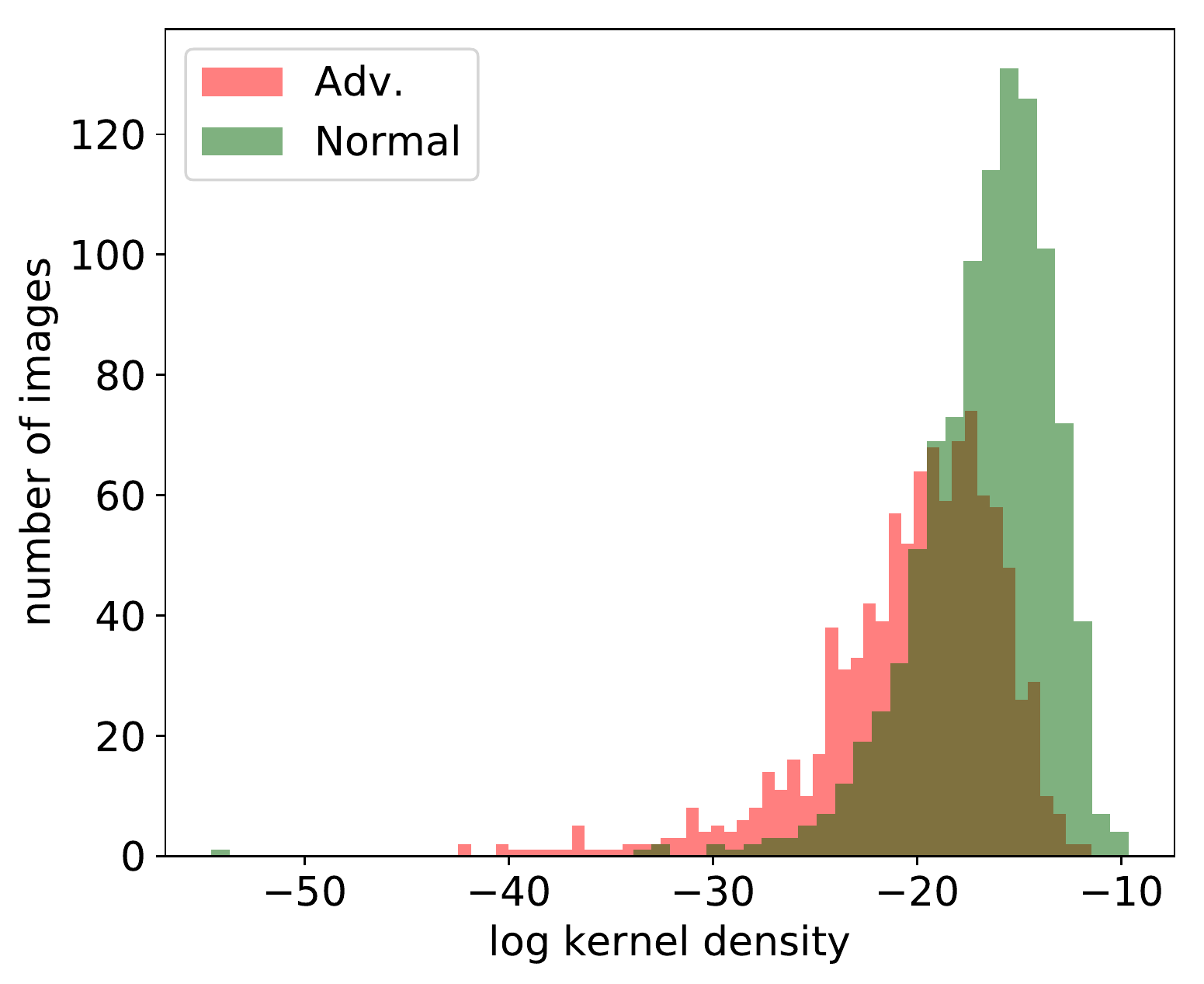}
	\end{minipage}}
	\subfigure{
		\begin{minipage}[t]{0.48\linewidth}
			\centering
			\includegraphics[width=1\linewidth]{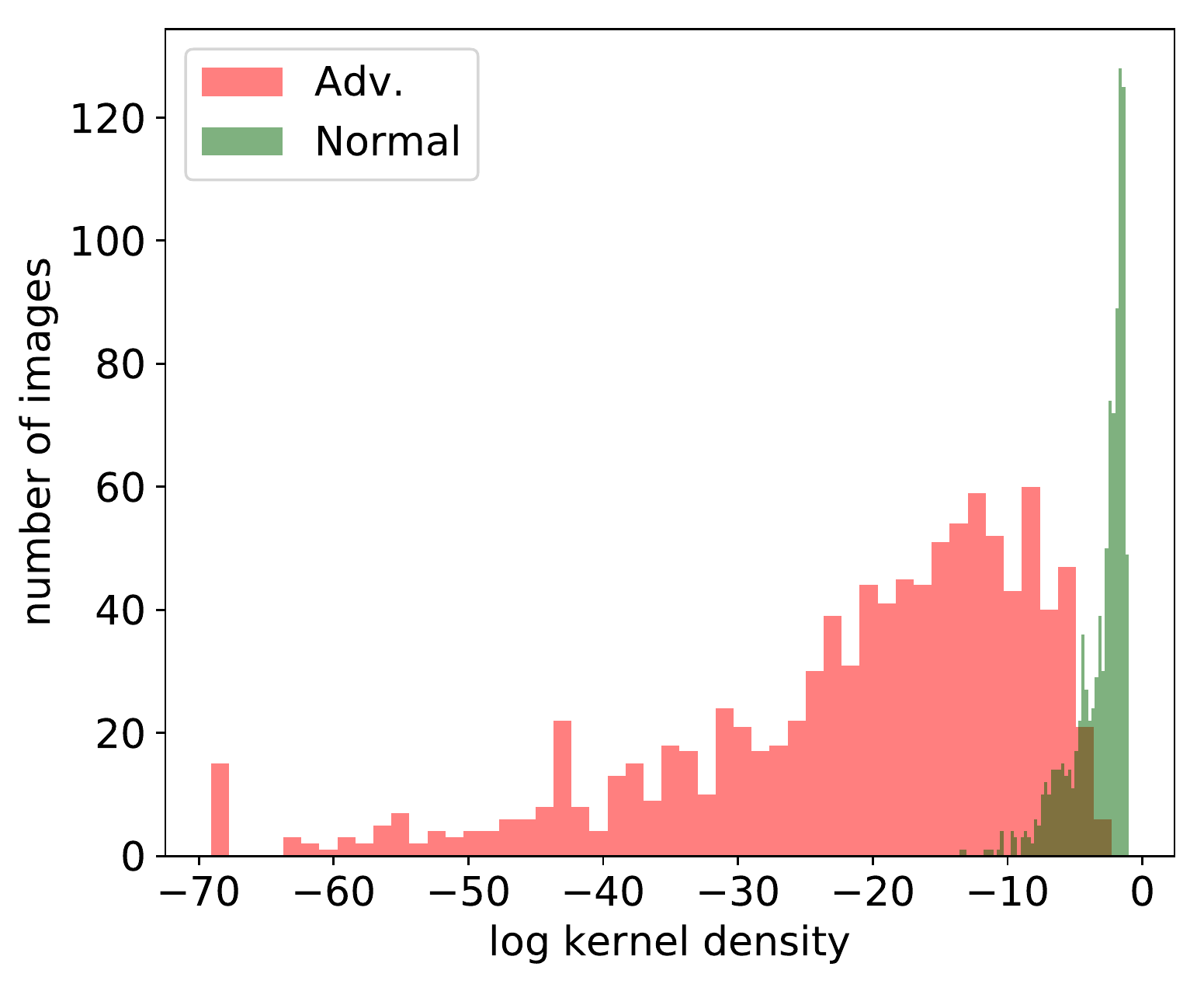}
	\end{minipage}}
	
	\subfigure{
		\begin{minipage}[t]{0.48\linewidth}
			\centering
			\includegraphics[width=1\textwidth]{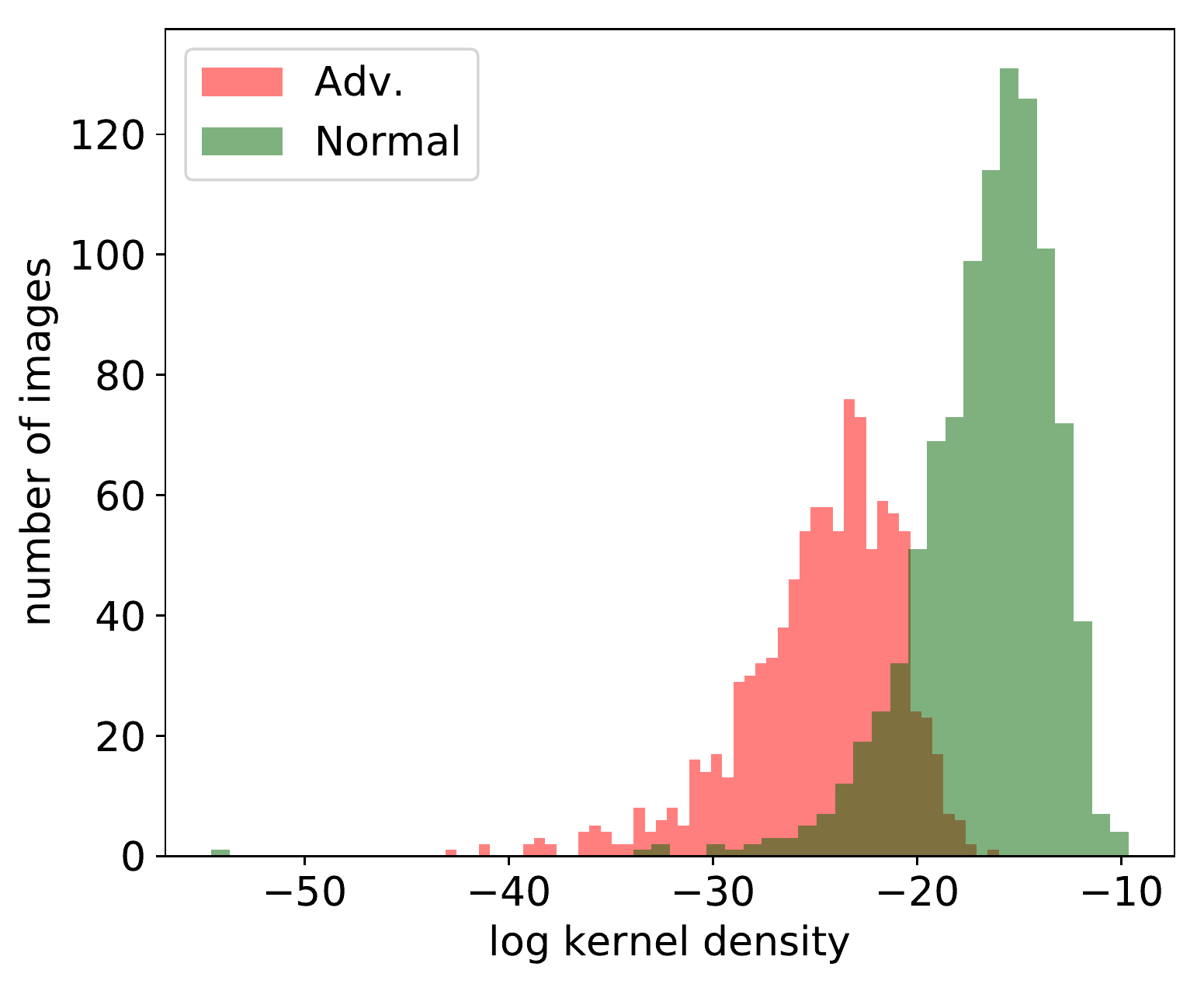}
	\end{minipage}}
	\subfigure{
		\begin{minipage}[t]{0.48\linewidth}
			\centering
			\includegraphics[width=1\textwidth]{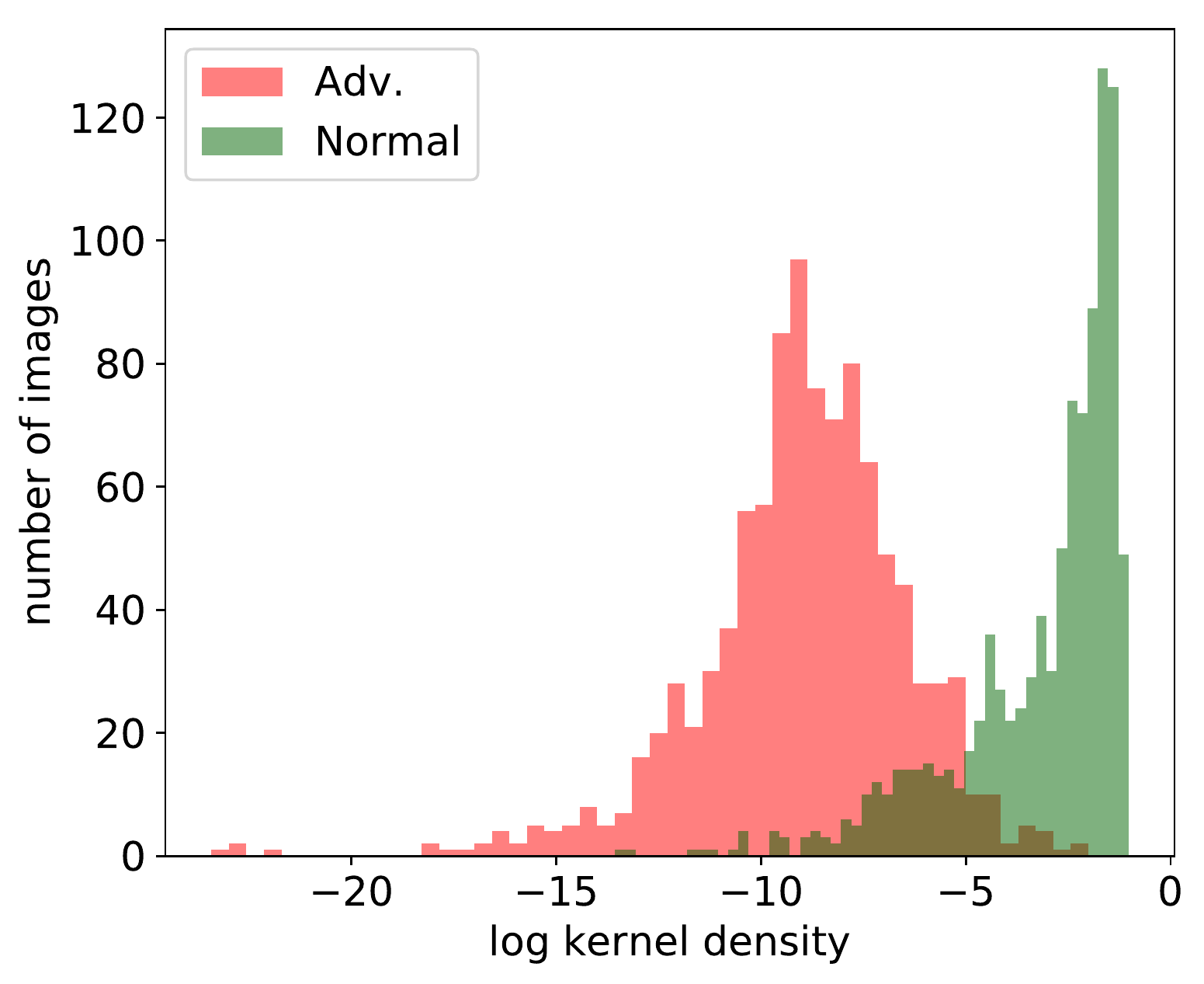}
	\end{minipage}}
	
	\centering
	\caption{\textbf{K-density histograms of adversarial examples generated on CIFAR-10 for models trained with different loss functions.} \textbf{Left column}: softmax loss; \textbf{Right column}: GM loss. From top to bottom, adversarial attack methods are FGSM~\cite{goodfellow2014explaining}, BIM~\cite{bim} and C\&W~\cite{carlini2017towards}.}
	\label{fig_hist}
\end{figure}

\begin{table}[t]
	\caption{\textbf{Detection AUC-ROC scores (\%) for semi white-box attacks on the MNIST and CIFAR datasets. }$\dagger$ indicates that this result is obtained by our re-implementation due to different early-stop settings~\cite{pang2018rce}.}
	\label{tb_semi_cifar}
	\centering
	\begin{tabular}{lc@{\hspace{1\tabcolsep}}c@{\hspace{1\tabcolsep}}c@{\hspace{1\tabcolsep}}c} 
		\Xhline{2\arrayrulewidth}
		Detection Method & FGSM~\cite{goodfellow2014explaining} & BIM~\cite{bim} & ILCM~\cite{bim} & C\&W~\cite{carlini2017towards}  \\ 
		\hline
		\midrule 
		\multicolumn{5}{l}{\textit{Dataset: MNIST}} \\ 
		Softmax+KD  &  90.2 & 90.3 & 95.8 & 97.6 \\
		KD+BU~\cite{feinman2017detecting} & 90.6 & 97.2 & 87.4 & 97.9  \\
		I-defender \textsubscript{NIPS'18}~\cite{zheng2018intrinsic} & 96.4 & 97.6 & - & -  \\
		LID\textsubscript{ICLR'18}~\cite{lid} & 96.9 & 99.6 & - & -  \\
		RCE \textsubscript{NIPS'18}~\cite{pang2018rce}  & 99.4 & 91.8 & 98.6 & 99.8  \\
		GM+KD(ours) & \textbf{99.5} & \textbf{99.7} & \textbf{99.0} & \textbf{99.8}  \\
		\hline 
		\multicolumn{5}{l}{\textit{Dataset: CIFAR-10}} \\
		Softmax+KD   & 71.2  & 60.2 &     85.4 & 76.5  \\
		KD+BU~\cite{feinman2017detecting} & 72.2 & 81.1 & 93.3 & 92.2  \\
		I-defender \textsubscript{NIPS'18}~\cite{zheng2018intrinsic} & 87.6 & 79.0 & - & -  \\
		LID\textsubscript{ICLR'18}~\cite{lid} & 82.4 & 82.5 & - & -  \\
		RCE \textsubscript{NIPS'18}~\cite{pang2018rce}  & 98.0 & 97.6$^\dagger$ & 93.9  & 98.2  \\
		GM+KD(ours) & \textbf{99.3} & \textbf{98.6} & \textbf{98.9} & \textbf{99.2}  \\
		\Xhline{2\arrayrulewidth}
	\end{tabular}
\end{table}

\subsubsection{Semi White-box Attack}
We evaluate the proposed method under the semi white-box setting.
Adversarial examples are generated based on the gradients of the pre-trained target model for all the normal images in the test sets.
The baseline method uses the K-density as a metric using the model trained with softmax loss.
The AUC-ROC scores are used for performance evaluation. 

For the MNIST and CIFAR-10 datasets, the results are summarized in Table~\ref{tb_semi_cifar}.
Overall, the proposed methods perform favorably against the baseline.
As an attack-agnostic approach, the proposed model can effectively detect all types of adversarial examples evaluated, especially the ILCM and C\&W-hc methods which are strong attacks on CIFAR-10.
The proposed method performs favorably against previous state-of-the-art approaches consistently for various attacks.
Furthermore, our method does not require extra training for other auxiliary modules as the I-defender does.
The baseline method performs well on the MNIST dataset.
As the MNIST dataset is a relatively simple dataset with small intra-class variations even in pixel-level, 
compact features of each class which are far from the decision boundary can be formed even when using the softmax loss.
For the more complex CIFAR-10 dataset, although with the same number of classes and a similar number of samples, other schemes under almost all attacks do not perform well. 
%
However, the proposed detection method generally maintains robust for both the MNIST and CIFAR-10 datasets.

\begin{table}[t]
	\caption{\textbf{Detection AUC-ROC scores (\%) for semi white-box attacks on the ImageNet dataset.}}
	\label{tb_semi_imagenet}
	\centering
	\begin{tabular}{lc@{\hspace{1\tabcolsep}}c@{\hspace{1\tabcolsep}}c@{\hspace{1\tabcolsep}}c} 
		\Xhline{2\arrayrulewidth}
		Detection Method & FGSM~\cite{goodfellow2014explaining} & BIM~\cite{bim} & ILCM~\cite{bim} & C\&W~\cite{carlini2017towards} \\ 
		\hline
		\hline
		Softmax+KD & 74.9 & 72.6 & 91.7 & 85.6 \\
		KD+BU~\cite{feinman2017detecting} & 90.6 & 87.2 & 92.1 & 94.8 \\
		GM+KD(ours) & \textbf{98.4} & \textbf{92.6} & \textbf{97.0} & \textbf{99.5} \\
		\Xhline{2\arrayrulewidth}
	\end{tabular}
\end{table}

The histograms of K-density for normal samples and adversarial examples are shown in Figure~\ref{fig_hist}.
The adversarial examples are generated under semi white-box settings.
The proposed model can better distinguish adversarial examples when combined with the K-density detector.

The proposed model is evaluated on a more challenging dataset, \ie ImageNet, which is less used for adversarial attacks in prior work.
Both RCE~\cite{pang2018rce} and proposed method are based on modeling the feature distribution.
However, this is a challenging task on the ImageNet as it has 100 times more classes than CIFAR-10 or MNIST datasets, with much larger intra-class variations.
%
For the ResNet-50 on the ImageNet, the training by using RCE loss encounters difficulty in convergence, which is probably caused by the overly strong constraint of forcing the networks to produce a uniform distribution on the classes other than the ground-truth class.
Therefore, we compare our method with the baseline and KD + Bayesian Uncertainty (BU)~\cite{feinman2017detecting}.
%
Table~\ref{tb_semi_imagenet} demonstrates that the proposed method can effectively detect adversarial examples on the ImageNet dataset.
%

\subsubsection{White-box attack}
\label{sec_white_box}
We evaluate our method under the white-box setting in which the adversarial examples are generated to fool the DNN classifier and the defense strategy simultaneously.
This setting is considered as the most difficult threat model.
In addition to the AUC-ROC scores, we can evaluate the defense methods using the average distortion of adversarial examples as a metric.
The distortion is defined as $\norm{x-x_{adv}}_2/\sqrt{hwc}$  \cite{szegedy2013intriguing}, which indicates the average $l_2$ distance for each pixel between the original image $x$ and the adversarial example $x_{adv}$.
The white-box variant of C\&W attack~\cite{carlini2017bypass} (denoted as C\&W-wb) is designed to attack the DNN classification model and K-density detector simultaneously.
The C\&W-wb attack introduces an extra loss term $f_2(x)=\max(-\log(KD(F(x)))-\eta,0)$, in which $\eta$ is the median of negative log K-density values on the training set.
We use the publicly available implementation released by the authors~\cite{carlini2017bypass} to perform this white-box attack.

\begin{table}[t]
	\caption{\textbf{Evaluation results on the CIFAR-10 dataset under white-box attacks for different methods.} Adversarial examples are generated using the C\&W-wb method~\cite{carlini2017bypass}.}
	\label{tb_white_CIFAR}
	\centering
	\begin{tabular}{lccc} 
		\Xhline{2\arrayrulewidth}
		Detection Method & AUC (\%)$\uparrow$ & Dist. $\uparrow$ & Adv. Success (\%)$\downarrow$ \cr
		\midrule
		Softmax+KD  & 53.8 & 1.26 & 100   \\
		RCE ~\cite{pang2018rce}  & 64.6 & 3.89 & 88.0 \\
		GM+KD (ours) & \textbf{84.3} & \textbf{7.82} & \textbf{70.1}  \\
		\Xhline{2\arrayrulewidth}
	\end{tabular}
\end{table}

\begin{table}[t]
	\caption{\textbf{Evaluation results on the MNIST dataset under white-box attacks for different methods.} Adversarial examples are generated using the C\&W-wb method~\cite{carlini2017bypass}.}
	\label{tb_white_MNIST}
	\centering
	\begin{tabular}{lcccccc} 
		\Xhline{2\arrayrulewidth}
		Detection Method & AUC (\%)$\uparrow$ & Dist. $\uparrow$ & Adv. Success (\%)$\downarrow$ \cr
		\midrule
		Softmax+KD & 55.6 & 18.42 & 98.8   \\
		RCE ~\cite{pang2018rce} & 67.9 & 31.59 & 86.9  \\
		GM+KD (ours)& \textbf{84.9} & \textbf{42.29} & \textbf{68.1}  \\
		\Xhline{2\arrayrulewidth}
	\end{tabular}
\end{table}

\renewcommand{\arraystretch}{1.2}
\begin{table}[t]
	\caption{\textbf{AUC-ROC socres (\%) for the black-box attacks for different DNN models.} Adversarial examples are generated using C\&W-wb method~\cite{carlini2017bypass}. The substitute and target models are ResNet-32 and ResNet-56, respectively.}
	\label{tb_black}
	\centering
	\begin{tabular}{cccc|ccc} 
		\Xhline{2\arrayrulewidth}
		Dataset & \multicolumn{3}{c}{MNIST} & \multicolumn{3}{c}{CIFAR-10} \\ 
		Loss Function & Softmax & RCE & GM & Softmax & RCE & GM \\
		\midrule
		AUC-ROC & 94.4 & 96.7 & \textbf{98.6} & 75.0 &84.9 & \textbf{94.7} \\
		
		\Xhline{2\arrayrulewidth}
	\end{tabular}
\end{table}

The results are presented in Table~\ref{tb_white_CIFAR} and \ref{tb_white_MNIST} where a success means that the adversarial example is classified into the target class and the extra loss $f_2(x) = 0$, and
the average success indicates the success rate of generating adversarial examples.
The white-box C\&W attacks are so strong that the baseline can hardly detect them since the AUC scores are nearly 50\% (random guess).
However, our methods achieves significantly higher AUC scores.
In addition, it is more difficult to generate such adversarial examples for our models which can be observed from both the high distortion and low adversarial success rate.

\subsubsection{Black-box Attack}
We evaluate the robustness of our methods for black-box attacks, in which the adversarial examples are generated from a substitute model and then used to attack the target model.
The attackers know the details of the defense strategy.
We employ the C\&W-wb method for evaluation. 
Table~\ref{tb_black} shows that
it is difficult for such adversarial examples to transfer from one model to another because of the discrepancy of feature distributions in the substitute and target models.
As a result, the proposed method can effectively detect such adversarial examples.

\section{Conclusions}
\label{conclusions}

Aiming to shape the deep feature space in image classification towards a Gaussian Mixture (GM) distribution, we propose a new loss function for training deep neural networks.
In addition to the classification loss which models the posterior probability distribution under the GM assumption, a novel likelihood regularization term is added to explicitly drive the deep features to follow the GM distribution.
We present in-depth analysis to reveal that theoretically, a symmetric feature space in which the distances between different class means are equal can be achieved when the proposed GM loss is optimized.
Such a feature space topology has been shown to effectively improve robustness against adversarial attacks.
To further improve the generalization capability of the trained model, a classification margin is introduced by slightly modifying the classification loss term.
The GM loss introduces no extra trainable parameters compared with the commonly used softmax loss.
Furthermore, the GM loss successfully boosts the performance of classification tasks and improves the robustness for adversarial attacks simultaneously.
Extensive experimental results demonstrate that the proposed GM loss performs favorably against the softmax loss and its variants on both small and large-scale datasets when combined with different deep models.
By introducing the kernel density estimation as the measurement for distinguishing adversarial examples from normal samples, the GM loss facilitates detecting adversarial examples robustly compared to previous state-of-the-art approaches. 
%


\ifCLASSOPTIONcaptionsoff
\newpage
\fi

\bibliographystyle{IEEEtran}
\bibliography{tpami19_GM_loss} 

\begin{thebibliography}{10}
\providecommand{\url}[1]{#1}
\csname url@samestyle\endcsname
\providecommand{\newblock}{\relax}
\providecommand{\bibinfo}[2]{#2}
\providecommand{\BIBentrySTDinterwordspacing}{\spaceskip=0pt\relax}
\providecommand{\BIBentryALTinterwordstretchfactor}{4}
\providecommand{\BIBentryALTinterwordspacing}{\spaceskip=\fontdimen2\font plus
\BIBentryALTinterwordstretchfactor\fontdimen3\font minus
  \fontdimen4\font\relax}
\providecommand{\BIBforeignlanguage}[2]{{%
\expandafter\ifx\csname l@#1\endcsname\relax
\typeout{** WARNING: IEEEtran.bst: No hyphenation pattern has been}%
\typeout{** loaded for the language `#1'. Using the pattern for}%
\typeout{** the default language instead.}%
\else
\language=\csname l@#1\endcsname
\fi
#2}}
\providecommand{\BIBdecl}{\relax}
\BIBdecl

\bibitem{DBLP:conf/eccv/WenZL016}
Y.~Wen, K.~Zhang, Z.~Li, and Y.~Qiao, ``A discriminative feature learning
  approach for deep face recognition,'' in \emph{European Conferenc on Computer
  Vision}, 2016.

\bibitem{DBLP:conf/icml/LiuWYY16}
W.~Liu, Y.~Wen, Z.~Yu, and M.~Yang, ``Large-margin softmax loss for
  convolutional neural networks,'' in \emph{International Conference on Machine
  Learning}, 2016.

\bibitem{goodfellow2014explaining}
I.~J. Goodfellow, J.~Shlens, and C.~Szegedy, ``Explaining and harnessing
  adversarial examples,'' \emph{arXiv preprint arXiv:1412.6572}, 2014.

\bibitem{DBLP:conf/nips/KrizhevskySH12}
A.~Krizhevsky, I.~Sutskever, and G.~E. Hinton, ``Imagenet classification with
  deep convolutional neural networks,'' in \emph{Annual Conference on Neural
  Information Processing Systems}, 2012.

\bibitem{DBLP:conf/icml/IoffeS15}
S.~Ioffe and C.~Szegedy, ``Batch normalization: Accelerating deep network
  training by reducing internal covariate shift,'' in \emph{International
  Conference on Machine Learning}, 2015.

\bibitem{DBLP:conf/cvpr/HeZRS16}
K.~He, X.~Zhang, S.~Ren, and J.~Sun, ``Deep residual learning for image
  recognition,'' in \emph{IEEE Conference on Computer Vision and Pattern
  Recognition}, 2016.

\bibitem{DBLP:conf/cvpr/SchroffKP15}
F.~Schroff, D.~Kalenichenko, and J.~Philbin, ``Facenet: A unified embedding for
  face recognition and clustering,'' in \emph{IEEE Conference on Computer
  Vision and Pattern Recognition}, 2015.

\bibitem{DBLP:conf/nips/SunCWT14}
Y.~Sun, Y.~Chen, X.~Wang, and X.~Tang, ``Deep learning face representation by
  joint identification-verification,'' in \emph{Annual Conference on Neural
  Information Processing Systems}, 2014.

\bibitem{Dahl2011Context}
D.~G. E., Y.~Dong, D.~Li, and A.~Alex, ``Context-dependent pretrained deep
  neural networks for large-vocabulary speech recognition,'' \emph{IEEE
  Transactions on Audio Speech and Language Processing}, vol.~20, no.~1, pp.
  30--42, 2011.

\bibitem{Hinton2012Deep}
H.~Geoffrey, D.~Li, Y.~Dong, D.~G. E., M.~Abdelrahman, J.~Navdeep, S.~Andrew,
  V.~Vincent, N.~Patrick, and S.~T. N., ``Deep neural networks for acoustic
  modeling in speech recognition: The shared views of four research groups,''
  \emph{IEEE Signal Processing Magazine}, vol.~29, no.~6, pp. 82--97, 2012.

\bibitem{DBLP:journals/corr/SzegedyVISW15}
C.~Szegedy, V.~Vanhoucke, S.~Ioffe, J.~Shlens, and Z.~Wojna, ``Rethinking the
  inception architecture for computer vision,'' \emph{arXiv preprint
  arXiv:1512.00567}, 2015.

\bibitem{zagoruyko2016wide}
S.~Zagoruyko and N.~Komodakis, ``Wide residual networks,'' \emph{arXiv preprint
  arXiv:1605.07146}, 2016.

\bibitem{larsson2016fractalnet}
G.~Larsson, M.~Maire, and G.~Shakhnarovich, ``Fractalnet: Ultra-deep neural
  networks without residuals,'' \emph{arXiv preprint arXiv:1605.07648}, 2016.

\bibitem{huang2017densely}
G.~Huang, Z.~Liu, L.~van~der Maaten, and K.~Q. Weinberger, ``Densely connected
  convolutional networks,'' in \emph{IEEE Conference on Computer Vision and
  Pattern Recognition}, 2017.

\bibitem{christopher2006pattern}
C.~M. Bishop, \emph{Pattern recognition and machine learning}.\hskip 1em plus
  0.5em minus 0.4em\relax Springer, 2006.

\bibitem{deng2014large}
J.~Deng, N.~Ding, Y.~Jia, A.~Frome, K.~Murphy, S.~Bengio, Y.~Li, H.~Neven, and
  H.~Adam, ``Large-scale object classification using label relation graphs,''
  in \emph{European Conference on Computer Vision}, 2014.

\bibitem{kobayashiLarge}
T.~Kobayashi, ``Large margin in softmax cross-entropy loss,'' in \emph{British
  Machine Vision Conference}, 2019.

\bibitem{feinman2017detecting}
R.~Feinman, R.~R. Curtin, S.~Shintre, and A.~B. Gardner, ``Detecting
  adversarial samples from artifacts,'' \emph{arXiv preprint arXiv:1703.00410},
  2017.

\bibitem{pang2018rce}
T.~Pang, C.~Du, Y.~Dong, and J.~Zhu, ``Towards robust detection of adversarial
  examples,'' in \emph{Advances in Neural Information Processing Systems},
  2018.

\bibitem{gm}
W.~Wan, Y.~Zhong, T.~Li, and J.~Chen, ``Rethinking feature distribution for
  loss functions in image classification,'' in \emph{IEEE Conference on
  Computer Vision and Pattern Recognition}, 2018.

\bibitem{pang2018max}
T.~Pang, C.~Du, and J.~Zhu, ``Max-mahalanobis linear discriminant analysis
  networks,'' in \emph{International Conference on Machine Learning}, 2018.

\bibitem{schroff2015facenet}
F.~Schroff, D.~Kalenichenko, and J.~Philbin, ``Facenet: A unified embedding for
  face recognition and clustering,'' in \emph{IEEE Conference on Computer
  Vision and Pattern Recognition}, 2015.

\bibitem{imagenet}
A.~Krizhevsky, I.~Sutskever, and G.~E. Hinton, ``Imagenet classification with
  deep convolutional neural networks,'' in \emph{Annual Conference on Neural
  Information Processing Systems}, 2012.

\bibitem{softtriple}
Q.~Qian, L.~Shang, B.~Sun, J.~Hu, H.~Li, and R.~Jin, ``Softtriple loss: Deep
  metric learning without triplet sampling,'' in \emph{IEEE international
  Conference on Computer Vision}, 2019.

\bibitem{Hinton2015Distilling}
H.~Geoffrey, V.~Oriol, and D.~Jeff, ``Distilling the knowledge in a neural
  network,'' \emph{arXiv preprint arXiv:1503.02531}, 2015.

\bibitem{Chen2017Noisy}
B.~Chen, W.~Deng, and J.~Du, ``Noisy softmax: Improving the generalization
  ability of dcnn via postponing the early softmax saturation,'' in \emph{IEEE
  Conference on Computer Vision and Pattern Recognition}, 2017.

\bibitem{szegedy2013intriguing}
C.~Szegedy, W.~Zaremba, I.~Sutskever, J.~Bruna, D.~Erhan, I.~Goodfellow, and
  R.~Fergus, ``Intriguing properties of neural networks,'' \emph{arXiv preprint
  arXiv:1312.6199}, 2013.

\bibitem{kurakin2016adversarial}
A.~Kurakin, I.~Goodfellow, and S.~Bengio, ``Adversarial machine learning at
  scale,'' \emph{arXiv preprint arXiv:1611.01236}, 2016.

\bibitem{madry2018towards}
A.~Madry, A.~Makelov, L.~Schmidt, D.~Tsipras, and A.~Vladu, ``Towards deep
  learning models resistant to adversarial attacks,'' in \emph{International
  Conference on Learning Representations}, 2018.

\bibitem{papernot2016distillation}
N.~Papernot, P.~McDaniel, X.~Wu, S.~Jha, and A.~Swami, ``Distillation as a
  defense to adversarial perturbations against deep neural networks,'' in
  \emph{2016 IEEE Symposium on Security and Privacy (SP)}, 2016.

\bibitem{carlini2017towards}
N.~Carlini and D.~Wagner, ``Towards evaluating the robustness of neural
  networks,'' in \emph{IEEE Symposium on Security and Privacy (SP)}, 2017.

\bibitem{metzen2017detecting}
J.~H. Metzen, T.~Genewein, V.~Fischer, and B.~Bischoff, ``On detecting
  adversarial perturbations,'' \emph{arXiv preprint arXiv:1702.04267}, 2017.

\bibitem{lu2017safetynet}
J.~Lu, T.~Issaranon, and D.~Forsyth, ``Safetynet: Detecting and rejecting
  adversarial examples robustly,'' in \emph{Proceedings of the IEEE
  International Conference on Computer Vision}, 2017.

\bibitem{defenseGAN}
P.~Samangouei, M.~Kabkab, and R.~Chellappa, ``Defense-gan: Protecting
  classifiers against adversarial attacks using generative models,''
  \emph{arXiv preprint arXiv:1805.06605}, 2018.

\bibitem{song2017pixeldefend}
Y.~Song, T.~Kim, S.~Nowozin, S.~Ermon, and N.~Kushman, ``Pixeldefend:
  Leveraging generative models to understand and defend against adversarial
  examples,'' \emph{arXiv preprint arXiv:1710.10766}, 2017.

\bibitem{zheng2018intrinsic}
Z.~Zheng and P.~Hong, ``Robust detection of adversarial attacks by modeling the
  intrinsic properties of deep neural networks,'' in \emph{Advances in Neural
  Information Processing Systems}, 2018.

\bibitem{durrieu2012lower}
J.-L. Durrieu, J.-P. Thiran, and F.~Kelly, ``Lower and upper bounds for
  approximation of the kullback-leibler divergence between gaussian mixture
  models,'' in \emph{IEEE International Conference on Acoustics, Speech and
  Signal Processing (ICASSP)}, 2012.

\bibitem{DBLP:conf/aaai/SunCWLL16}
S.~Sun, W.~Chen, L.~Wang, X.~Liu, and T.~Liu, ``On the depth of deep neural
  networks: A theoretical view,'' in \emph{AAAI Conference on Artificial
  Intelligence}, 2016.

\bibitem{tensorflow}
\BIBentryALTinterwordspacing
M.~Abadi \emph{et~al.}, ``{TensorFlow}: Large-scale machine learning on
  heterogeneous systems,'' 2015, software available from tensorflow.org.
  [Online]. Available: \url{https://www.tensorflow.org/}
\BIBentrySTDinterwordspacing

\bibitem{mnist}
Y.~LeCun, L.~Bottou, Y.~Bengio, and P.~Haffner, ``Gradient-based learning
  applied to document recognition,'' \emph{Proceedings of the IEEE}, vol.~86,
  no.~11, pp. 2278--2324, 1998.

\bibitem{krizhevsky2009learning}
A.~Krizhevsky and G.~Hinton, ``Learning multiple layers of features from tiny
  images,'' University of Toronto, Tech. Rep., 2009.

\bibitem{sutskever2013importance}
I.~Sutskever, J.~Martens, G.~Dahl, and G.~Hinton, ``On the importance of
  initialization and momentum in deep learning,'' in \emph{International
  Conference on Machine Learning}, 2013.

\bibitem{he2015delving}
K.~He, X.~Zhang, S.~Ren, and J.~Sun, ``Delving deep into rectifiers: Surpassing
  human-level performance on imagenet classification,'' in \emph{IEEE
  international Conference on Computer Vision}, 2015.

\bibitem{DBLP:journals/corr/SimonyanZ14a}
K.~Simonyan and A.~Zisserman, ``Very deep convolutional networks for
  large-scale image recognition,'' \emph{arXiv preprint arXiv:1409.1556}, 2014.

\bibitem{DBLP:journals/corr/IoffeS15}
S.~Ioffe and C.~Szegedy, ``Batch normalization: Accelerating deep network
  training by reducing internal covariate shift,'' \emph{arXiv preprint
  arXiv:1502.03167}, 2015.

\bibitem{DBLP:journals/corr/LinCY13}
M.~Lin, Q.~Chen, and S.~Yan, ``Network in network,'' \emph{arXiv preprint
  arXiv:1312.4400}, 2013.

\bibitem{deng2009imagenet}
J.~Deng, W.~Dong, R.~Socher, L.-J. Li, K.~Li, and L.~Fei-Fei, ``Imagenet: A
  large-scale hierarchical image database,'' in \emph{IEEE Conference on
  Computer Vision and Pattern Recognition}, 2009.

\bibitem{dropout}
G.~E. Hinton, N.~Srivastava, A.~Krizhevsky, I.~Sutskever, and R.~R.
  Salakhutdinov, ``Improving neural networks by preventing co-adaptation of
  feature detectors,'' \emph{arXiv preprint arXiv:1207.0580}, 2012.

\bibitem{resnet}
K.~He, X.~Zhang, S.~Ren, and J.~Sun, ``Deep residual learning for image
  recognition,'' in \emph{IEEE Conference on Computer Vision and Pattern
  Recognition}, 2016.

\bibitem{xie2017aggregated}
S.~Xie, R.~Girshick, P.~Doll{\'a}r, Z.~Tu, and K.~He, ``Aggregated residual
  transformations for deep neural networks,'' in \emph{Proceedings of the IEEE
  conference on computer vision and pattern recognition}, 2017.

\bibitem{senet}
J.~Hu, L.~Shen, and G.~Sun, ``Squeeze-and-excitation networks,'' in \emph{IEEE
  Conference on Computer Vision and Pattern Recognition}, 2018.

\bibitem{huang2007labeled}
G.~B. Huang, M.~Ramesh, T.~Berg, and E.~Learned-Miller, ``Labeled faces in the
  wild: A database for studying face recognition in unconstrained
  environments,'' University of Massachusetts, Amherst, Tech. Rep., 2007.

\bibitem{DBLP:journals/corr/YiLLL14a}
D.~Yi, Z.~Lei, S.~Liao, and S.~Z. Li, ``Learning face representation from
  scratch,'' \emph{arXiv preprint arXiv:1411.7923}, 2014.

\bibitem{zhang2016joint}
K.~Zhang, Z.~Zhang, Z.~Li, and Y.~Qiao, ``Joint face detection and alignment
  using multitask cascaded convolutional networks,'' \emph{IEEE Signal
  Processing Letters}, vol.~23, no.~10, pp. 1499--1503, 2016.

\bibitem{sun2015deeply}
Y.~Sun, X.~Wang, and X.~Tang, ``Deeply learned face representations are sparse,
  selective, and robust,'' in \emph{IEEE Conference on Computer Vision and
  Pattern Recognition}, 2015.

\bibitem{Liu2014Deep}
Z.~Liu, P.~Luo, X.~Wang, and X.~Tang, ``Deep learning face attributes in the
  wild,'' in \emph{International Conference on Computer Vision}, 2014.

\bibitem{cifar}
A.~Krizhevsky and G.~Hinton, ``Learning multiple layers of features from tiny
  images,'' University of Toronto, Tech. Rep., 2009.

\bibitem{bim}
A.~Kurakin, I.~Goodfellow, and S.~Bengio, ``Adversarial examples in the
  physical world,'' \emph{arXiv preprint arXiv:1607.02533}, 2016.

\bibitem{adam}
D.~P. Kingma and J.~Ba, ``Adam: A method for stochastic optimization,''
  \emph{arXiv preprint arXiv:1412.6980}, 2014.

\bibitem{lid}
X.~Ma, B.~Li, Y.~Wang, S.~M. Erfani, S.~Wijewickrema, G.~Schoenebeck, M.~E.
  Houle, D.~Song, and J.~Bailey, ``Characterizing adversarial subspaces using
  local intrinsic dimensionality,'' in \emph{International Conference on
  Learning Representations}, 2018.

\bibitem{carlini2017bypass}
N.~Carlini and D.~Wagner, ``Adversarial examples are not easily detected:
  Bypassing ten detection methods,'' in \emph{Proceedings of the 10th ACM
  Workshop on Artificial Intelligence and Security}, 2017.

\end{thebibliography}

\begin{IEEEbiography}[{\includegraphics[width=1in,height=1.25in,clip,keepaspectratio]{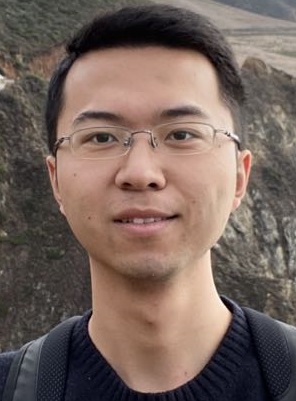}}]{Weitao Wan} is a Ph.D. candidate of in the Department of Electronic Engineering, Tsinghua University, Beijing, China.
	He received the B.S. degree in Electronic Engineering from Tsinghua University, Beijing, China, in 2016.
	His research interests include computer vision, adversarial learning and weakly-supervised learning.
\end{IEEEbiography}

\begin{IEEEbiography}[{\includegraphics[width=1in,height=1.25in,clip,keepaspectratio]{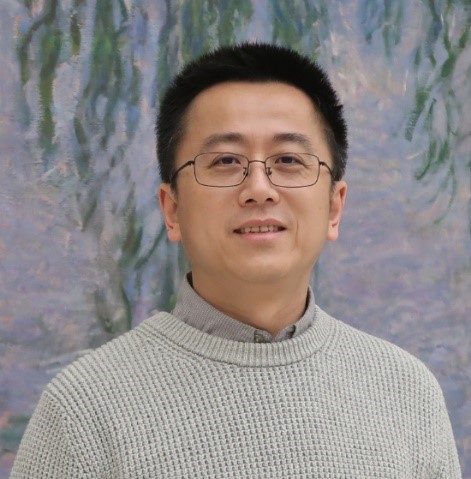}}]{Jiansheng Chen} is an associate professor in the Department of Electronic Engineering, Tsinghua University, Beijing, China. He received the B.E. and M.E. degrees, both in computer science and technology, from Tsinghua University, Beijing, China, in 2000 and 2002, respectively. He received the Ph.D. degree
	in computer science and engineering from the Chinese University of Hong Kong in 2007.  His research interests include image processing, pattern recognition and machine learning.
\end{IEEEbiography}

\begin{IEEEbiography}[{\includegraphics[width=1in,height=1.25in,clip,keepaspectratio]{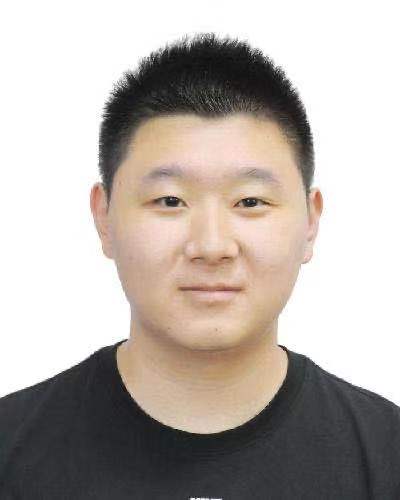}}]{Cheng Yu}
	is a Ph.D student in the Department of Electronic Engineering, Tsinghua University, Beijing, China. He received the B.S. degree in Electronic 
	Engineering from Tsinghua University, Beijing, China, in 2018.
	His research interests include computer vision, machine learning, and adversarial learning.
\end{IEEEbiography}

\begin{IEEEbiography}[{\includegraphics[width=1in,height=1.25in,clip,keepaspectratio]{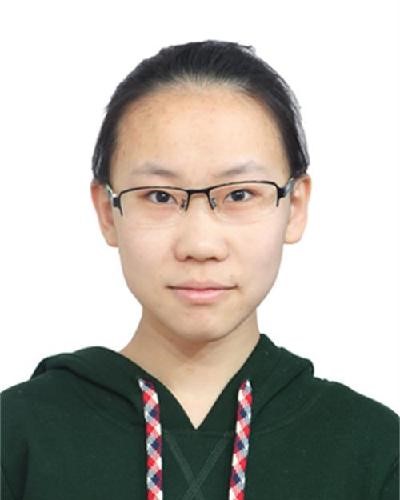}}]{Tong Wu}
	Tong Wu is currently a senior student and she has been attended in Tsinghua University since 2016. She will receive the B.S. degree in electronic engineering from Tsinghua University, Beijing, China in 2020. She is currently working on her graduation project. Her research experiences and research interests include computer vision, deep learning and machine learning.
\end{IEEEbiography}

\begin{IEEEbiography}[{\includegraphics[width=1in,height=1.25in,clip,keepaspectratio]{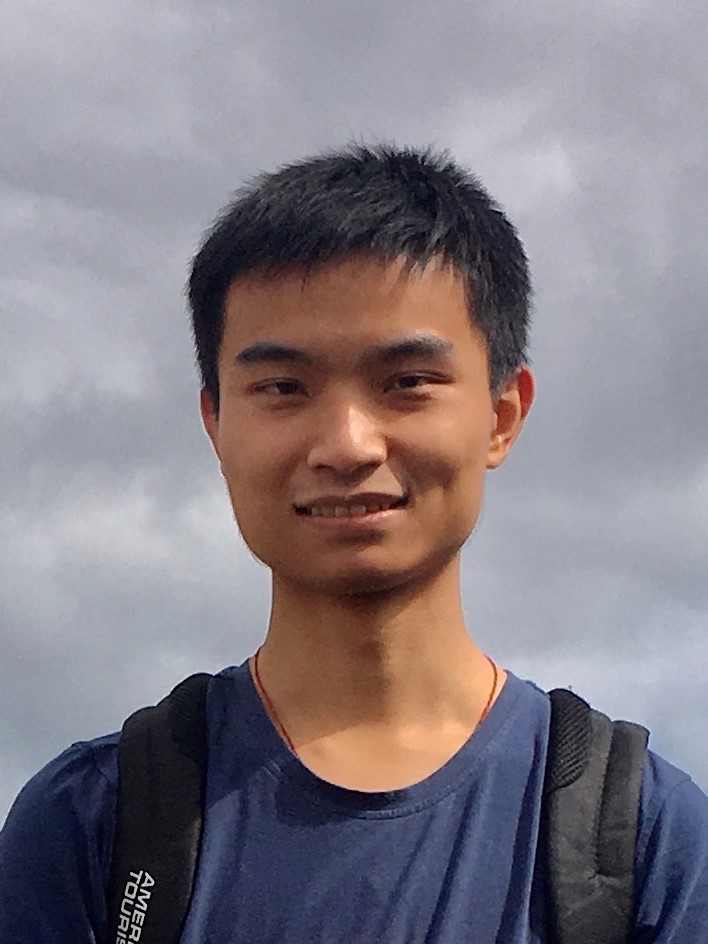}}]{Yuanyi Zhong}
	Yuanyi Zhong received his B.E. in Electronic Engineering from Tsinghua University in 2017. He is currently pursuing a Ph.D. degree in the Department of Computer Science, University of Illinois at Urbana Champaign, IL, USA. His research interests include machine learning, computer vision (object recognition) and reinforcement learning, especially with the use of deep neural networks.
\end{IEEEbiography}

\begin{IEEEbiography}[{\includegraphics[width=1in,height=1.25in,clip,keepaspectratio]{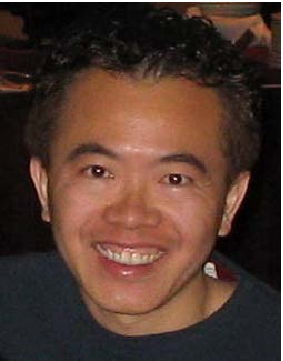}}]{Ming-Hsuan Yang}
	is a professor in Electrical
	Engineering and Computer Science at University
	of California, Merced. 
	Yang serves as a program co-chair of IEEE International Conference on Computer Vision (ICCV) in 2019, program co-chair of Asian Conference on Computer Vision (ACCV) in 2014, and general co-chair of ACCV 2016.
	Yang served as an associate editor of the IEEE
	Transactions on Pattern Analysis and Machine
	Intelligence from 2007 to 2011, and is an associate
	editor of the International Journal of Computer
	Vision, Image and Vision Computing and
	Journal of Artificial Intelligence Research. He
	received the NSF CAREER award in 2012
	and Google Faculty Award in 2009. 
	He is a Fellow of the IEEE.
\end{IEEEbiography}

\end{document}